\documentclass{article}

\usepackage[main, final, nonatbib]{neurips_2026}

\usepackage[utf8]{inputenc} %
\usepackage[T1]{fontenc}    %
\usepackage{hyperref}       %
\usepackage{url}            %
\usepackage{booktabs}       %
\usepackage{tabularx}       %
\usepackage{amsfonts}       %
\usepackage{microtype}      %
\usepackage{xcolor}         %
\usepackage{amsmath}        %
\usepackage{graphicx}       %
\usepackage{placeins}       %
\usepackage{enumitem}       %
\usepackage{wrapfig}        %
\usepackage{needspace}       %
\graphicspath{{figures/}}

\newif\ifcameraready
\camerareadytrue  %

\usepackage[style=numeric-comp, url=false,isbn=false,doi=false, eprint=false,giveninits=true,maxbibnames=99,maxcitenames=2,date=year,sorting=none,sortcites=true, natbib=true]{biblatex}
\DeclareSourcemap{
  \maps[datatype=bibtex]{
    \map{
      \step[fieldsource=doi,final]
      \step[fieldset=url,null]
    }
  }
}
\title{Decomposing Earth Embeddings with Sparse~Autoencoders}%

\author{%
  Vitus Benson$^{1,2,3}$\thanks{Correspondence: \texttt{vbenson@bgc-jena.mpg.de}} \quad
  Fanny Yang$^{3}$ \quad
  Markus Reichstein$^{1,2}$ \\[0.4em]
  $^{1}$Max Planck Institute for Biogeochemistry, Jena, Germany \\
  $^{2}$ELLIS Unit Jena, Jena, Germany \\
  $^{3}$ETH Zürich, Zürich, Switzerland
}

\begin{document}

\maketitle

\begin{abstract}
Earth observation foundation models produce global metre-resolution embeddings that underpin work across agriculture, ecology, hydrology, and beyond.
Yet these embeddings are opaque: each dimension entangles unrelated physical concepts, which makes them difficult to interpret and their robustness hard to assess.
Here, we propose sparse autoencoders (SAEs) as a principled way to decompose Earth embeddings into sparse, approximately monosemantic features over an overcomplete dictionary.
Because the standard practice of using LLMs to judge interpretations does not transfer to satellite pixels, and spatial autocorrelation matters, we introduce a tailored evaluation protocol.
It scores any SAE on reconstruction at matched sparsity, alignment with reference concepts, and spatial scale, each against baselines that include an SAE whose decoder is a frozen random dictionary.
On this protocol, SAEs reconstruct AlphaEarth and Tessera embeddings better than every baseline at matched sparsity.
Their features align with far fewer reference concepts than raw embedding dimensions (on AlphaEarth, $88\%$ align with at most one of $20$ concept groups, most with none, against a median of $12$ per raw dimension).
They also carry more of their variance at fine spatial scales ($23$ to $32\%$ within a $640$\,m patch for BatchTopK and Matryoshka, against $9\%$).
An SAE with a frozen random decoder shares both properties, which therefore follow from sparsity, whereas only a learned dictionary reconstructs the embedding accurately: on AlphaEarth, trained SAEs leave eight times less variance unexplained.
We test the sparse code on two applications, discovering land-cover sub-concepts and detecting out-of-distribution pixels by area-of-applicability estimation.
Both indicate how sparse features could be used, but neither is yet reliable enough for practical use.
Together, these results are a step toward Earth observation models whose features can be inspected, opening pathways for scientific discovery and more trustworthy geospatial applications.
Code and the dataset recipe are released at \url{https://github.com/EarthyScience/geosae}.
\end{abstract}

\section{Introduction}\label{sec:intro}

Earth embeddings compress petabytes of satellite imagery into compact, reusable vector representations that capture the state of the land surface at any location and time~\cite{klemmer.etal_2025}.
Precomputed embeddings such as AlphaEarth~\cite{brown.etal_2025a} and Tessera~\cite{feng.etal_2025b}, and open models such as OlmoEarth~\cite{herzog.etal_2025b}, already enable rapid global mapping from sparse labels.
They have been used successfully in agriculture~\cite{zvonkov.etal_2025, fang.etal_2026, ma.etal_2026}, forestry~\cite{tsao.lobell_2025}, meteorology~\cite{giezendanner.etal_2026}, hydrology~\cite{qu.etal_2026}, development economics~\cite{pettersson.daoud_2025}, disaster risk reduction~\cite{feng.etal_2026}, and environmental monitoring~\cite{houriez.etal_2025}. For the first time, the full Earth observation archives are within practical reach for downstream analysis.

These embeddings, however, remain largely opaque.
Their intrinsic dimensionality is far lower than their ambient size~\cite{rao.etal_2026}, which suggests substantial redundancy.
Some dimensions correlate with specific land-surface properties, but most mix several concepts~\cite{benavides-martinez.etal_2026, rahman_2026}.
This entanglement undermines two properties that practitioners need.
The first is \emph{interpretability}, since scientists and decision makers must understand what a model has learned to trust its predictions and to discover new insights.
The second is \emph{robustness under distribution shift}.
Downstream models are often trained on labels from one region and applied globally.
If a dimension encodes deciduous forest in the training region but irrigated cropland elsewhere, the model transfers its forest mapping to cropland.
Cropland pixels may then lie close to forest training pixels in embedding space, so a distance-based check for input shift~\cite{meyer.pebesma_2022a} can miss the error~\cite{ma.etal_2026, corley.etal_2026a}.
If forest and cropland activated different features, such a shift would become visible.

Sparse autoencoders (SAEs) offer a principled approach to this problem.
The use of sparse, overcomplete representations has a long history in computational neuroscience~\cite{olshausen.field_1996} and has recently emerged as the dominant tool for mechanistic interpretability.
By projecting dense activations onto an overcomplete dictionary, SAEs disentangle representations into approximately monosemantic features, each corresponding to a single interpretable concept~\cite{bricken2023monosemanticity, huben.etal_2023}.
Originally developed for language models~\cite{gao.etal_2024b, bussmann.etal_2024b, bussmann.etal_2025b, karvonen.etal_2025, korznikov.etal_2025}, SAEs have since been applied to protein language models~\cite{gujral.etal_2025}, single-cell transcriptomics~\cite{claye.etal_2025}, biological foundation models more broadly~\cite{orlov.etal_2026}, vision transformers~\cite{stevens.etal_2025, stevens.etal_2025a}, weather and climate models~\cite{macmillan.ouellette_2025, king.etal_2025}, and physics surrogates~\cite{hu.etal_2026}, consistently surfacing interpretable structure that remains invisible in the original dense activations of these large-scale foundation models.
We bring this approach to Earth embeddings.

\begin{figure}[!t]
    \centering
    \includegraphics[width=\linewidth]{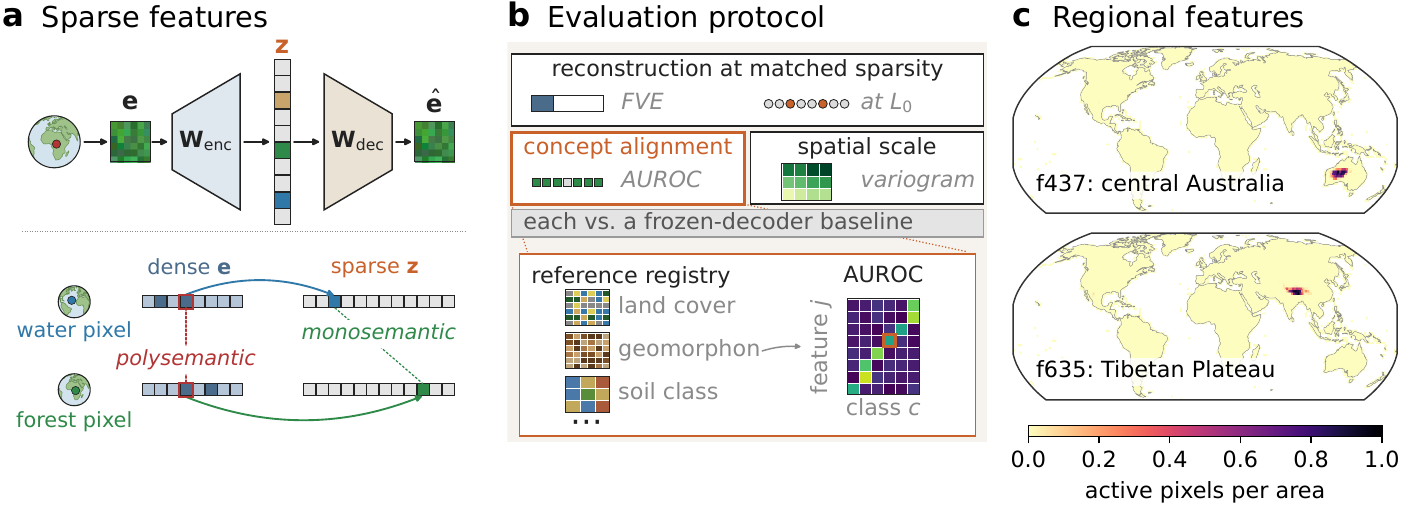}
    \caption{Overview. (a)~An SAE decomposes the embedding of each pixel into a sparse code whose features are meant to be monosemantic, whereas a dense dimension is polysemantic.
    (b)~The evaluation protocol scores reconstruction at matched sparsity, alignment with reference concepts and spatial scale, each compared with baselines, including an SAE with a frozen random decoder.
    (c)~For two SAE features, the share of pixels on which the feature is active in each $2^\circ{\times}2^\circ$ cell (about $220$\,km), averaged over the test patches in the cell.}
    \label{fig:overview}
\end{figure}

The main obstacle to using SAEs on Earth embeddings is evaluation.
Current SAE evaluation protocols were developed for language models and commonly rely on LLMs to interpret features automatically, which is not established for Earth observation.
An LLM can read the text on which a language-model feature fires, but the meaning of a pixel has to be read from maps of what is on the ground.
In Earth science, moreover, concepts are spatially autocorrelated~\cite{tobler_1970, rolf.etal_2024} and vary on scales from single fields to biomes, so the scale at which a feature varies is part of what it represents.
A further difficulty concerns SAEs in general: on language models, SAEs whose decoder is a frozen random dictionary nearly match trained ones on common interpretability metrics~\cite{korznikov.etal_2026}, and SAEs on randomly initialised transformers score about as well as SAEs on trained ones~\cite{heap.etal_2025}.
A good score alone therefore does not show that the SAE has learned meaningful features.
We therefore propose an evaluation protocol tailored to Earth embeddings (Figure~\ref{fig:overview}b).
It scores any SAE on three criteria: reconstruction at matched sparsity, alignment with known concepts, and spatial scale.
The known concepts come from a curated registry of reference maps, ranging from sub-metre roads to $500$\,m land-cover grids, and every score is compared with that of an SAE with a frozen random decoder.

\Needspace{7\baselineskip}
Our contributions are as follows:
\begin{enumerate}
    \item We apply sparse autoencoders to pixelwise Earth embeddings for the first time, decomposing AlphaEarth and Tessera embeddings into sparse codes over overcomplete dictionaries (Figure~\ref{fig:overview}a).
    We propose an evaluation protocol for these SAEs with three criteria, each compared with baselines that include a frozen-decoder SAE (Figure~\ref{fig:overview}b).
    \item We compare three state-of-the-art SAE variants (ReLU+$\ell_1$, BatchTopK, Matryoshka BatchTopK) with PCA, ICA, NMF and the frozen-decoder baseline.
    All three reconstruct the embeddings better than every baseline at matched sparsity.
    Their features align with far fewer concept groups than raw dimensions and vary on finer spatial scales.
    The frozen-decoder baseline shares both properties, which therefore follow from sparsity, whereas training improves reconstruction quality.
    \item We test the sparse code on two applications: discovering sub-concepts inside coarse land-cover classes, and area-of-applicability estimation for out-of-distribution (OOD) detection.
    Both hint at how sparse features could be used, but neither is yet reliable enough for practical use.
\end{enumerate}

In concurrent work, Betti et al.~\cite{betti.etal_2026} apply SAEs to location encoders, which map geographic coordinates to embeddings.
In another concurrent work, Young and Keshav~\cite{young.keshav_2026} train one SAE per region on AlphaEarth and Tessera pixels and interpret it over time, through features that track disturbances such as fires.
We instead train global SAEs on pixelwise embeddings and evaluate them with a protocol that applies to any SAE, including the two monosemanticity scores of Betti et al.\ (Appendix~\ref{app:betti_scores}). 

\section{Setup}\label{sec:setup}

We train SAEs on the embeddings of two geospatial foundation models, which represent individual $10$\,m pixels of satellite imagery with a vector (§\ref{sec:embeddings}).
We introduce the SAE variants (§\ref{sec:sae_primer}), our evaluation protocol with its baselines (§\ref{sec:interpretability}), the two applications (§\ref{sec:downstream}) and the data (§\ref{sec:data}).

\subsection{Earth embeddings}\label{sec:embeddings}

We use the term \emph{Earth embedding} for any learned vector representation of a location on Earth~\cite{klemmer.etal_2025, fang.etal_2026a}.
It is produced either by a location encoder from geographic coordinates or by an image encoder from satellite imagery, as one vector per patch or per pixel.
Here, we work with pixelwise embeddings, in which a geospatial foundation model represents every $10$\,m pixel of the satellite imagery in a given year by a vector $\mathbf{e}\in\mathbb{R}^D$.
We use two publicly available datasets of such embeddings, each released once per year.

\paragraph{AlphaEarth Foundations}~\cite{brown.etal_2025a} is the model that produced the Google Satellite Embedding dataset, whose $D{=}64$ embeddings we use.
It treats the imagery of a location as a video: an encoder with spatial and temporal self-attention summarises all observations of a year into one unit-norm embedding per $10$\,m pixel.
During training, decoders reconstruct from this summary the trajectories of many sources over the same period: optical multispectral imagery (Sentinel-2 and Landsat), synthetic-aperture radar (Sentinel-1), GEDI spaceborne lidar, MODIS land-surface temperature, ERA5 climate reanalysis, digital elevation and label maps; the embedding is also aligned with text.
To produce the released embeddings, however, the trained encoder is applied to Sentinel-1/2 and Landsat imagery only.
Because the encoder attends over space, the embedding of a pixel also depends on the pixels around it.

\paragraph{Tessera}~\cite{feng.etal_2025b} computes $D{=}128$ embeddings from the time series of each single pixel over one calendar year, without looking at neighbouring pixels.
Two transformer branches encode the Sentinel-2 reflectances (10 bands) and the Sentinel-1 radar backscatter (2 polarisations), and a small network fuses both into one embedding.
The model was trained with Barlow Twins~\cite{zbontar.etal_2021}: two random subsets of the cloud-free observation dates of the same pixel are pushed towards the same embedding, while its dimensions are decorrelated.
This objective helps to learn a pixel's annual trajectory regardless of a clear-sky sampling bias.
Tessera does not explicitly consider spatial relationships, so any spatial structure in the embeddings instead must come from the underlying imagery.

\subsection{Sparse autoencoders}\label{sec:sae_primer}

A \emph{sparse autoencoder} (SAE) maps each dense vector $\mathbf{e}\in\mathbb{R}^D$ to a much wider latent vector, its \emph{code} $\mathbf{z}\in\mathbb{R}^M$ ($M\gg D$) with only a few non-zero entries, and reconstructs $\mathbf{e}$ from it:
\begin{equation}
    \mathbf{z} = \mathcal{S}\!\left(\mathbf{W}_{\mathrm{enc}}\,\mathbf{e}+\mathbf{b}_\text{enc}\right), \qquad \hat{\mathbf{e}} = \mathbf{W}_{\mathrm{dec}}\,\mathbf{z} + \mathbf{b}, \qquad \mathcal{L}_\text{rec} = \|\mathbf{e}-\hat{\mathbf{e}}\|_2^2.
    \label{eq:sae}
\end{equation}
Here $\mathbf{W}_{\mathrm{enc}}\in\mathbb{R}^{M\times D}$ and $\mathbf{W}_{\mathrm{dec}}\in\mathbb{R}^{D\times M}$ are the encoder and decoder weights, $\mathbf{b}_\text{enc}$ and $\mathbf{b}$ are biases (Appendix~\ref{sec:supp_training} gives the parametrisation used in training), and the sparsifier $\mathcal{S}$ sets most entries of $\mathbf{z}$ to zero.
Encoder and decoder are trained to minimise the reconstruction error $\mathcal{L}_\text{rec}$.
The columns of $\mathbf{W}_{\mathrm{dec}}$, the decoder directions, are the \emph{atoms} of a dictionary in the sense of sparse coding~\cite{olshausen.field_1996}; they are renormalised to unit norm before every training step.
The dictionary is \emph{overcomplete}: it holds more atoms than the embedding has dimensions.
The entry $z_j$, the activation of atom $j$, is \emph{feature} $j$ (also called a latent); over all pixels it forms a map, and the feature is active where $z_j\neq 0$.
Because only a few features may be active for any input, each is pressured to specialise on a single recurring pattern of the data.
The design goal is that each feature be approximately \emph{monosemantic}~\cite{bricken2023monosemanticity}, responding to one concept that a human can name and staying near zero otherwise; a \emph{polysemantic} feature responds to several unrelated concepts.

We benchmark three SAE variants that differ in how they enforce sparsity.
\textbf{ReLU+$\ell_1$}~\cite{huben.etal_2023} uses $\mathcal{S}=\mathrm{ReLU}$ and adds an $\ell_1$-penalty on $\mathbf{z}$ to the loss, which encourages sparsity without prescribing a fixed level.
\textbf{BatchTopK}~\cite{bussmann.etal_2024b} keeps the $B{\cdot}k$ largest pre-activations across a training batch of $B$ pixels, so that on average $k$ features are active per pixel; it relaxes TopK~\cite{gao.etal_2024b}, which keeps exactly $k$ per pixel.
At test time a fixed threshold, estimated during training, replaces this selection, so each pixel is encoded on its own.
\textbf{Matryoshka BatchTopK}~\cite{bussmann.etal_2025b} adds a loss term in which the first $n_s$ atoms alone must reconstruct $\mathbf{e}$, which encourages a coarse-to-fine organisation of concepts (Eq.~\ref{eq:matryoshka}). We set $n_s = \mathrm{round}(0.2\,M)$ in our experiments.
Both BatchTopK variants add an auxiliary loss that revives dead features, i.e.\ those that have stopped being active on any pixel~\cite{gao.etal_2024b}.
We train each variant with $M{=}1024$ atoms and $k\in\{8,16,32\}$, and with $M\in\{512,2048\}$ at $k{=}16$, three seeds each; for ReLU+$\ell_1$, the penalty weight is tuned so that the mean number of active features matches $k$.
We evaluate reconstruction on all these models, and spatial scale and concept alignment on one reference model per variant, with $M{=}1024$ and $k{=}16$.
Appendix~\ref{sec:supp_training} contains all training details.

\subsection{Evaluation}\label{sec:interpretability}
SAEs for language models are commonly evaluated by automated interpretability.
An LLM describes a feature from the text on which it is active, and the description is then scored by how well it predicts the activations of the feature on new text~\cite{karvonen.etal_2025}.
A feature of an Earth-embedding SAE activates on pixels instead, for which no comparable protocol exists to judge interpretations.
We therefore introduce three criteria for evaluating SAEs for Earth embeddings, each motivated below.

\begin{enumerate}[leftmargin=1.4em]
\item \textbf{Reconstruction at matched sparsity.}
An interpretable representation should explain most of the embedding $\mathbf{e}$ in as few active features per pixel as possible.
In other words, at matched sparsity, its reconstruction $\hat{\mathbf{e}}$ should have lower error.
We measure sparsity by $L_0$, the mean number of active features per pixel.
For the SAEs, sparsity is a design choice, set by $k$ for BatchTopK and by the penalty weight for ReLU+$\ell_1$.
For the embeddings $\mathbf{e}_1,\dots,\mathbf{e}_N$ of $N$ test pixels (Appendix~\ref{sec:supp_eval}), we report the mean-squared error per embedding dimension, $\mathrm{MSE} = \tfrac{1}{ND}\sum_{n=1}^{N}\|\mathbf{e}_n-\hat{\mathbf{e}}_n\|_2^2$, and compare at matched $L_0$.
We also report the fraction of variance explained (FVE): one minus the summed squared reconstruction error, divided by the summed squared deviation of the embeddings from their mean.
Table~\ref{tab:eight_criteria} also reports the cosine similarity between embedding and reconstruction.
Note that the reconstruction criterion alone can be improved by increasing the expressivity of the decoder, i.e.\ by storing information in its weights instead of in the sparse latents.
While this is less of an issue in this work because we only consider linear decoders, it is still important to consider other evaluation criteria beyond reconstruction quality.

\item \textbf{Spatial scale.} Many geographic concepts vary smoothly in space (biomes, climate zones), whereas fence rows, roads and single trees vary at high spatial frequency, so we ask at which length scales each feature varies.
For feature $j$ with activation $z_j(s)$ at pixel location $s$, we estimate its semivariance at a separation of $h$ metres from the set $P_h$ of pixel pairs in the evaluation sample that lie $h$ apart, in any direction:
\begin{equation}
    \gamma_j(h) \;=\; \frac{1}{2\,|P_h|}\sum_{(s,s')\in P_h}\big(z_j(s)-z_j(s')\big)^2.
    \label{eq:variogram}
\end{equation}
We divide it by the variance $\sigma_j^2$ of the feature over all pixels of the sample.
This normalised semivariance rises from the \emph{nugget} $\gamma_j(10\,\mathrm{m})/\sigma_j^2$, the variation between neighbouring pixels, towards $1$ at large separations.
Within one $64{\times}64$-pixel patch of our dataset (§\ref{sec:data}), pixel pairs are $10$ to $630$\,m apart.
Pixel pairs from different patches cover separations from $13$\,km to the antipodes.
Because our patches are sampled sparsely, separations between $640$\,m and $13$\,km are not covered.
The \emph{correlation length} $\ell_j=\min\{h:\gamma_j(h)/\sigma_j^2\geq 1-e^{-1}\}$ is the smallest separation at which the normalised semivariance reaches $1-e^{-1}\approx0.63$, that is, at which the correlation $1-\gamma_j(h)/\sigma_j^2$ between activations $h$ apart has fallen to $1/e$.
A feature with a short correlation length changes over short distances, for instance from one field to the next, whereas a feature with a long correlation length is nearly constant within a region and changes only between distant places.
Its \emph{within-patch share} is the share of its variance at separations below $640$\,m.

\item \textbf{Concept alignment.} A useful Earth-embedding SAE should recover at least some of the geographic concepts that the Earth-science community already maps, such as land-cover types.
Published categorical maps, such as land-cover maps, assign each pixel one class, for example ``tree cover'' or ``cropland''.
We call these maps \emph{reference products} and take their classes as ground truth for concepts (§\ref{sec:data}).
To test whether feature $j$ represents class $c$, we use its activation $z_j$ as a score for membership in $c$.
On a sample balanced between pixels of class $c$ and other pixels, we compute the area under the ROC curve (AUROC) $A$ of this score.
We fold it, $\max(A, 1{-}A)$, so that a feature that marks a class by staying silent on it also counts, and call feature $j$ \emph{aligned} with $c$ if the folded AUROC is at least $0.65$.
This threshold trades genuine predictive signal against the noise in the labels.
The products are themselves imperfect and partly contradict each other, so we group similar classes into $20$ \emph{concept groups} (e.g.\ ``water'', ``cropland''; Table~\ref{tab:supp_concept_groups}); a feature is aligned with a group if it is aligned with at least one of its classes.

For each feature, we count the concept groups it is aligned with: one for a monosemantic feature, several for a polysemantic one, and none for a feature that matches no group.
This count is an approximate measure of monosemanticity.
For an SAE, we report the distribution of this count over its features, for example the share of features aligned with at most one group.
We count only \emph{live} features, those whose AUROC exceeds $0.51$ for at least one class; the others separate no class better than chance.

\end{enumerate}

\paragraph{Baselines.}\label{sec:controls}
In sparse coding, random dictionaries are known to perform well already when only the encoder is trained~\cite{coates.ng_2011}.
This insight has been shown to hold also for SAEs of language models, where a frozen random decoder can match trained SAEs on interpretability metrics~\cite{korznikov.etal_2026}.
Such an SAE however has not learned a dictionary and is hence uninterpretable.
We therefore compare the scores of our SAEs with those of a BatchTopK SAE of the same dictionary size $M$ and sparsity whose decoder is fixed to random unit-norm atoms and whose encoder alone is trained, the \emph{frozen-decoder baseline}.
This is a sanity check in the sense of Adebayo et al.~\cite{adebayo.etal_2018}, comparing an interpretation method on learned vs.\ random weights.

Other baseline feature representations that we compare with include the raw embedding and three standard decompositions of the embedding: principal component analysis (PCA), independent component analysis (ICA) and non-negative matrix factorisation (NMF), whose non-negative coefficients mirror the activations of an SAE.
Each of these decompositions has $D$ components, as many as the embedding has dimensions.
For reconstruction, each baseline uses only $L_0$ components (PCA its $L_0$ leading components, ICA and NMF $L_0$ fitted components, and the raw embedding its $L_0$ dimensions of largest variance), so that all methods are compared at matched sparsity (§\ref{sec:results_table}).
Using the baselines, we can examine whether a change of basis alone already yields interpretable axes, or whether an overcomplete, sparse dictionary helps (Appendix~\ref{sec:supp_baselines}).

\subsection{Applications}\label{sec:downstream}

Beyond these scores, we test the sparse code on two applications, one for interpretation and one for robust mapping.

\paragraph{Sub-concept discovery.} Fine-resolution global land-cover maps typically carry few classes, such as only a single ``forest'', ``water'' or ``built-up'' class, although domain experts distinguish many sub-types within each.
An SAE is not told at which level of such a hierarchy its features should sit, so several of its features can each respond to part of one coarse class, an effect known as feature splitting~\cite{bricken2023monosemanticity, chanin.etal_2025}.
Such features may mark sub-types.
We explore this phenomenon for woody vegetation with three hand-picked features.
For each feature, we take the $2{,}000$ test patches with the highest mean activation and look up the climate zone, biome and potential natural vegetation of each patch in three maps outside our registry: K\"oppen-Geiger climate zones~\cite{karger.etal_2017, peel.etal_2007}, RESOLVE biomes~\cite{dinerstein.etal_2017} and potential natural vegetation~\cite{hengl.etal_2018} (Appendix~\ref{sec:supp_subconcepts}).
We inspect the features at two scales.
Inside a single patch, we show the activation of a feature, which can then be compared with corresponding satellite imagery or reference products.
Worldwide, we divide the globe into cells of $2^\circ{\times}2^\circ$ (about $220$\,km at the equator) and map, for each cell, the share of pixels on which the feature is active, averaged over the test patches in the cell.

\paragraph{Area of applicability for OOD detection.}
Models trained on regional ground truth are routinely applied globally, so detecting the out-of-distribution (OOD) pixels where they extrapolate is a prerequisite for trustworthy use~\cite{meyer.pebesma_2021, meyer.pebesma_2022a}.
The area-of-applicability method of Meyer and Pebesma~\cite{meyer.pebesma_2021} flags a test pixel as OOD if it lies far from its nearest training pixels in the representation space of the model, with each dimension weighted by its importance for the prediction.
This distance can mislead if a dimension mixes unrelated concepts, because two genuinely unrelated pixels can then still lie close together in that dimension.
As a purely conceptual illustration, if one dimension encodes both water and trees, a canopy-height model trained on forest pixels may appear confident over water.
SAEs in contrast may help with approximately monosemantic features.
On natural images, sparse SAE codes detect OOD inputs with fewer false positives than methods based on dense activations~\cite{oh.etal_2026}.
Here we study computing the distance for the area-of-applicability estimates in the $M$-dimensional SAE code.
We test this approach in a synthetic setting with known ground truth: we fit a ridge regressor on a training set of pixels, and then predict on held-out pixels under covariate shift.
As targets we try canopy height (CHMv2~\cite{brandt.etal_2026}) and elevation (Copernicus DEM~\cite{copernicus_dem_2021}), and we test $88$ hold-outs per target that lie increasingly far from the training data in space or in climate.
For each held-out pixel, we compute the mean Euclidean distance, in the representation space, to its five nearest training pixels, either unweighted or with each dimension weighted by its absolute ridge coefficient.
A representation space scores well if this distance ranks the absolute prediction errors on the held-out pixels, which we measure by the Spearman rank correlation $\rho$ between the two.
We repeat the experiment using the raw embedding (conventional setting) and the SAE latents.
Appendix~\ref{sec:supp_aoa} provides further details.

\subsection{Data}\label{sec:data}

A single global map of Earth embeddings can easily exceed $100$\,TB at $10$\,m resolution.
For this reason, we instead rely on a random sample of $210{,}000$ patches of $64{\times}64$ pixels ($640{\times}640$\,m).
We draw the sample from the about $5.4$ million land cells of the MajorTOM grid~\cite{francis.czerkawski_2024}, whose $5$\,km spacing leaves gaps of about $4$\,km between adjacent patches.
For each patch we download the AlphaEarth and Tessera embeddings of 2024.
We split the patches into $100{,}000$ training, about $10{,}000$ validation and $100{,}000$ test patches, and report all numbers on the test patches.

\paragraph{Reference products.}
For interpretability evaluation, we compare SAE latents with reference maps.
We compile a registry of \textbf{14 categorical} and \textbf{3 continuous} reference products, with native resolutions from sub-metre roads to $500$\,m land-cover grids (Figure~\ref{fig:supp_dataset_overview} shows them on one patch; details in Appendix~\ref{sec:supp_data}).
Every product is resampled onto the $10$\,m grid of each patch, categorical products by nearest neighbour and continuous ones bilinearly.
Note that the products are themselves imperfect and refer to different years, for instance 2021 for ESA WorldCover against 2024 for the embeddings.
Single-feature alignment is scored against the categorical products.
In addition, the individual classes of all products are grouped into concept groups, which allow for testing alignment independently of any single product (§\ref{sec:interpretability}).
Table~\ref{tab:supp_data_sources} lists the products and their sources.

\section{Results}\label{sec:results}
We now report the three criteria described in §\ref{sec:interpretability} for the SAEs and all baselines (§\ref{sec:results_table} to §\ref{sec:results_emf}), and then discuss two applications of the sparse code (§\ref{sec:results_qualitative} and §\ref{sec:results_aoa}).
In each subsection we compare with the frozen-decoder baseline, investigating whether a property depends on the learned dictionary.
We present additional evidence for each subsection in Appendices~\ref{sec:supp_recon} to~\ref{sec:supp_aoa_results}, in the same order.

\begin{figure}[!t]
    \centering
    \includegraphics[width=\linewidth]{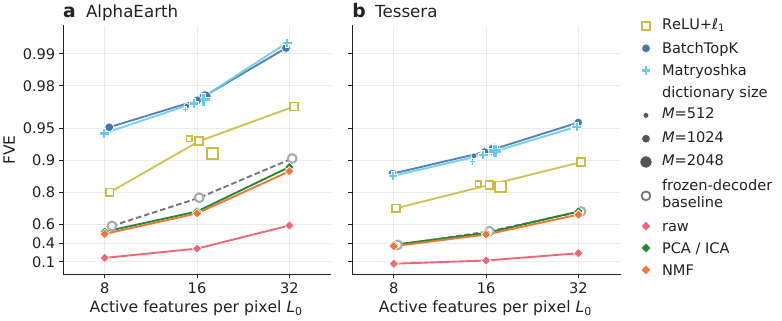}
    \caption{Reconstruction against sparsity on (a)~AlphaEarth and (b)~Tessera. The fraction of variance explained (FVE) against the mean number of active features per pixel, $L_0$. Marker size shows the dictionary size $M$. All trained SAEs reconstruct better than every baseline at every $L_0$.}
    \label{fig:pareto}
\end{figure}

\subsection{SAEs reconstruct better than every baseline}\label{sec:results_table}
Figure~\ref{fig:pareto} shows the fraction of variance explained (FVE) against the mean number of active features per pixel, $L_0$, for every SAE of the sweep and for the baselines with $L_0$ components.
At every $L_0$ and on both embeddings, all three SAE variants reconstruct better than every baseline.
At $L_0{=}16$, the reconstructions of BatchTopK and Matryoshka have a mean cosine similarity of $0.99$ with the AlphaEarth embedding and of $0.95$ to $0.96$ with the Tessera embedding, and those of ReLU+$\ell_1$ of $0.97$ and $0.92$.
The reconstruction quality depends mainly on $L_0$: changing $M$ from $512$ to $2048$ at $L_0{=}16$, or changing the training seed, moves FVE by less than $0.01$.

\begin{wrapfigure}[16]{r}{0.47\linewidth}
    \vspace{-1.1em}
    \centering
    \includegraphics[width=\linewidth]{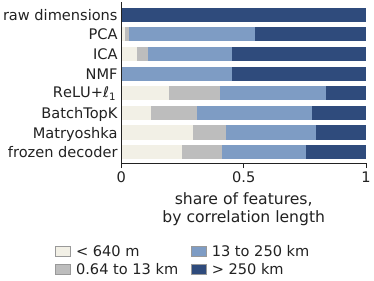}
    \caption{Share of features by correlation length $\ell_j$ (§\ref{sec:interpretability}). SAE features vary on finer scales than raw dimensions. Appendix~\ref{sec:supp_variogram_tessera} shows the full semivariance curves.}
    \label{fig:variogram}
    \vspace{-0.3em}
\end{wrapfigure}
The frozen-decoder baseline separates the two sources of this advantage.
Because it chooses which $16$ of its $1024$ random atoms to use, it already explains more of AlphaEarth than the first $16$ principal components ($0.78$ against $0.70$).
Learning the atoms then leaves significantly less variance unexplained, at every $L_0$ from $8$ to $32$ (at $L_0{=}16$, $2.7\%$ against $22\%$).
On Tessera the baseline explains about as much as PCA ($0.54$ against $0.52$), so the whole advantage of the SAEs there comes from training (at $L_0{=}16$, $8.4\%$ against $47\%$ unexplained).

\subsection{SAE features vary on finer scales}\label{sec:results_freq}
Figure~\ref{fig:variogram} shows, for each decomposition of AlphaEarth, the share of features in each range of correlation length.
The median correlation length is $1{,}240$\,km for a raw dimension and $200$ to $270$\,km for a PCA, ICA or NMF component: they change mainly between distant places.
In contrast, the median SAE feature has a correlation length of $20$ to $33$\,km, and the median share of a feature's variance within a patch is $23$ to $32\%$, against $9\%$ for a raw dimension.
Fine-scale variation is thus present in the embedding, but in a dense dimension it is outweighed by larger-scale variation, whereas a sparse code represents it with dedicated features.
Importantly, the shorter correlation lengths do not necessarily mean that the SAE features are fundamentally different.
In Appendix~\ref{sec:supp_dominant} we show that the leading SAE features are effectively discretised versions of the leading principal components of the embedding, each covering only one region of the world and thus having a shorter correlation length.

The frozen-decoder baseline spreads its features over scales similarly to the trained SAEs (median $23$\,km), so this spread does not depend on the learned dictionary.
The three SAE variants do differ slightly in their within-patch correlation structure, with ReLU+$\ell_1$ and BatchTopK having a much smaller median nugget than Matryoshka ($0.013$ and $0.016$, against $0.035$).
The same pattern arises when repeating the analysis with Tessera embeddings or using Moran's $I$ instead, a standard index of spatial autocorrelation (Appendix~\ref{sec:supp_spatial}).

\begin{figure}[!t]
    \centering
    \includegraphics[width=\linewidth]{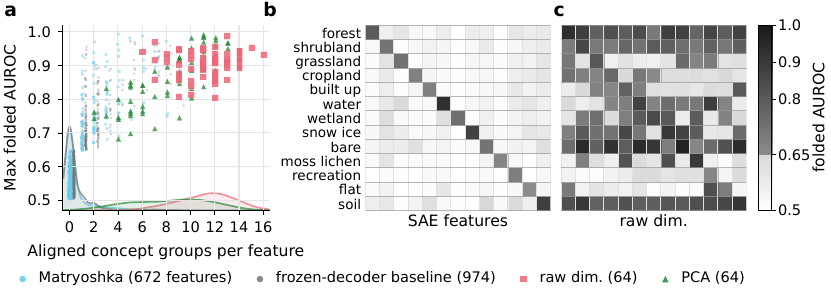}
    \caption{Concept alignment on AlphaEarth. (a)~Number of concept groups features are aligned with, against their largest folded AUROC; the legend gives the number of live features per model. (b,~c)~Folded AUROC of the best-aligned feature (column) for each concept group (row), for the Matryoshka SAE and the raw dimensions ($13$ of $20$ groups shown); dark cells are aligned.}
    \label{fig:monosemanticity}
\end{figure}

\subsection{SAE features align with fewer concepts than raw dimensions}\label{sec:results_emf}

We measure how monosemantic the features are by the number of concept groups each feature is aligned with (§\ref{sec:interpretability}).
Figure~\ref{fig:monosemanticity}a shows this number for every live feature of each representation of AlphaEarth, against the folded AUROC of the class it predicts best.
A raw dimension is aligned with a median of $12$ of the $20$ concept groups and a PCA component with $9$, so rotating the embedding does not remove the polysemanticity.
In contrast, $88\%$ of the $672$ live Matryoshka features are aligned with at most one group: $8\%$ with exactly one and $80\%$ with none.

Figures~\ref{fig:monosemanticity}b and~c show, for each concept group, the most specific feature aligned with it and how well this feature separates every other group (Appendix~\ref{sec:supp_alignment} describes the selection rule in detail).
The selected SAE features are aligned with their own group only, whereas the selected raw dimensions are still aligned with a median of $8$ groups each.
Yet the selected raw dimensions separate their own group better than the selected SAE features (mean folded AUROC $0.84$ against $0.75$): an SAE feature is more specific to a concept, but not a better detector of it.

The frozen-decoder baseline scores as well on this count: $89\%$ of its $974$ live features are aligned with at most one group, and it has more aligned features than the Matryoshka SAE ($234$ against $136$).
The reason is sparsity.
A feature that is zero on most pixels reaches a folded AUROC of $0.65$ for a group only if it is active on at least $30\%$ of the group's pixels (or of the other pixels), whereas with $16$ of $1024$ features active per pixel, a feature is active on $1.6\%$ of pixels on average, so most features are aligned with no group and few with several.
The count therefore describes how specific the features of a sparse code are, not what its dictionary has learned.
For two other scores that assess the monosemanticity, the trained SAEs do slightly outperform the frozen-decoder baseline: the number of product classes that some feature is aligned with ($140$ against $129$ at a threshold of $0.60$, one run each) and the mean monosemanticity scores of Betti et al.\ \cite{betti.etal_2026}.
Results are largely consistent for Tessera and BatchTopK (Appendix~\ref{sec:supp_alignment}).

\begin{figure}[!t]
    \centering
    \includegraphics[width=\linewidth]{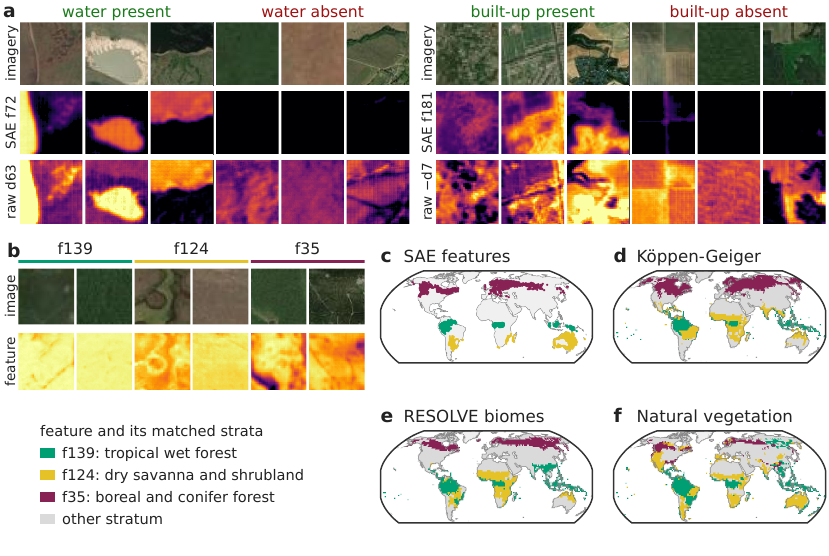}
    \caption{Aligned features on AlphaEarth. (a)~Imagery and activation of the best-aligned Matryoshka feature and raw dimension, on three test patches containing the class and three without ($-$ marks a sign-flipped dimension). (b)~Top-activating patches of three woody-vegetation features, (c)~their footprints and (d--f)~the matching strata of K\"oppen-Geiger climate zones, RESOLVE biomes and potential natural vegetation. Appendix~\ref{sec:supp_subconcepts} contains more examples.}
    \label{fig:qualitative}
\end{figure}

\subsection{Aligned features follow concepts within patches and across biomes}\label{sec:results_qualitative}
Beyond scores, we now qualitatively inspect learned SAE features against reference products.
Figure~\ref{fig:qualitative}a shows an example for two concepts, water and built-up land, each tracked by an SAE feature.
In contrast, the best-aligned raw dimension is clearly polysemantic.

We further investigate if the SAE features could lead to sub-concept discovery, where multiple SAE features describe sub-types of the same coarse class.
Figure~\ref{fig:qualitative}b--f shows three hand-picked Matryoshka SAE features that are active mainly on woody vegetation.
The $2{,}000$ most active patches of each feature are concentrated: $80\%$ of those of f139 lie in tropical rainforest climate, $39\%$ of those of f124 in hot semi-arid climate, and $91\%$ of those of f35 in subarctic and continental climates.
The footprints largely follow the matching strata of K\"oppen-Geiger climate zones~\cite{karger.etal_2017, peel.etal_2007}, RESOLVE biomes~\cite{dinerstein.etal_2017} and potential natural vegetation~\cite{hengl.etal_2018} (panels d to f).
The frozen-decoder baseline has such features as well, but of arguably lower quality (Appendix~\ref{sec:supp_subconcepts_frozen}).

Such sub-types are what an SAE can offer a domain expert: candidate divisions of a coarse class, found without labels.
Validating them needs care, because climate and biome maps vary smoothly in space: random compact subdivisions of forest patches of the same size agree with them about as well (Appendix~\ref{sec:supp_subconcepts}).
The three features are also regional, active mainly in one part of the world (Appendix~\ref{sec:supp_dominant}), so field data would have to show whether they separate vegetation types or mainly regions.

\subsection{SAE features can improve OOD detection}\label{sec:results_aoa}

\begin{wrapfigure}[17]{r}{0.45\linewidth}
    \vspace{-1.2em}
    \centering
    \includegraphics[width=\linewidth]{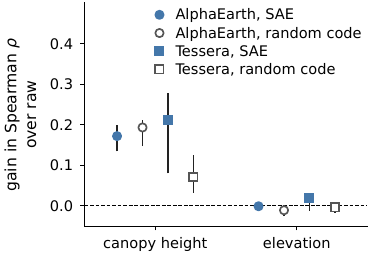}
    \caption{Gain in the rank correlation between area-of-applicability estimates and OOD prediction errors for SAE vs.\ raw embeddings; on $88$ hold-outs per target (median and $95\%$ bootstrap interval).}
    \label{fig:aoa_main}
    \vspace{-0.5em}
\end{wrapfigure}

Figure~\ref{fig:aoa_main} shows how much better the distance in SAE representation space ranks the errors of a downstream model than the distance in the raw embedding space (§\ref{sec:downstream}).
For canopy height, the ridge-weighted distance in a BatchTopK code ranks the errors better than the raw distance for most hold-outs and both embeddings ($77$ and $75$ of $88$ hold-outs, median gain in $\rho$ of $+0.17$ and $+0.21$).
For elevation, however, the SAE does not help.
This is not surprising: SAE-based OOD detection should help only in particular situations, for instance when the downstream model relies on polysemantic dimensions that mislead it outside the training region.
Even for canopy height, the learned dictionary is not what helps on AlphaEarth: an untrained random sparse code of the same size and sparsity gains as much as the SAE there, and about a third as much on Tessera ($+0.07$).
Moreover, the SAE-based estimate is not uniformly stricter than the raw one.
On the leave-Europe-out fold for canopy height, it marks $526$ of $6{,}910$ held-out pixels as outside the area of applicability that the raw estimate keeps, but it also keeps $271$ that the raw estimate marks as outside (Figure~\ref{fig:supp_aoa_diff}).
Even where such an estimate ranks the errors better, a practitioner who follows a precautionary principle may prefer the stricter one.

SAE-based applicability estimates are therefore a research idea rather than a ready tool.
First, they can also hurt: the SAE distance ranks the errors no better than the raw distance in $11$ and $13$ of the $88$ canopy-height hold-outs on AlphaEarth and Tessera, and in $47$ and $39$ of the $88$ elevation hold-outs.
Second, a distance in representation space measures only how unlike the training inputs a test pixel is.
It therefore targets \emph{covariate shift}, a change in the distribution of the inputs.
It cannot detect \emph{concept shift}, a change in how the target relates to the inputs, and it sees \emph{label shift}, a change in the distribution of the target, only through the inputs that come with it; both need labels from the new region.
Third, our hold-outs are synthetic: they are cut from one global sample in which the target is known everywhere, whereas a real transfer can combine all three kinds of shift.
Even in this controlled setting, whether the SAE helps depends on the target, so an estimate has to be checked against held-out errors for each new target before it is trusted.
Appendix~\ref{sec:supp_aoa_results} provides more results and discusses the implications in further detail.

\Needspace{6\baselineskip}
\section{Discussion}
Sparse autoencoders disentangle Earth embeddings into approximately monosemantic geographic features.
Our experiments show this consistently across two foundation models (AlphaEarth, Tessera) and three state-of-the-art SAE variants (ReLU+$\ell_1$, BatchTopK, Matryoshka BatchTopK).
Our evaluation protocol scores any SAE on reconstruction, spatial scale and concept alignment.
Comparing with a frozen-decoder baseline, an SAE whose random decoder is never trained, separates what sparsity alone provides from what learning the dictionary adds.
Sparsity alone makes features specific and spreads them over spatial scales from single pixels to continents, whereas only a learned dictionary reconstructs the embedding accurately: with $16$ active features, trained SAEs leave eight times less of the variance of AlphaEarth unexplained.
Interpretability scores of single features should therefore be reported next to such a baseline, as for language models~\cite{korznikov.etal_2026, heap.etal_2025}.
We also find first evidence that SAE features may support two applications.
Their regional nature allows us to identify sub-concepts of coarse reference classes, entirely without labels.
We demonstrate this application with the example of dividing woody vegetation into tropical, dry-savanna and boreal.
Under specific circumstances, they can also give a more reliable area-of-applicability estimate than the raw embedding for a downstream model transferred to a new region, which we show for synthetic covariate shift scenarios based on canopy height prediction, although in other cases, the SAE can also hurt.

While these applications show the practical promise of SAEs in Earth observation, we view this work as a first step toward new opportunities for scientific discovery.
Once SAE architectures tailored to geospatial data learn dictionaries that capture regional and local structure, scientists could attribute sensitivities of Earth system processes to regional SAE-derived concepts.
For instance, describing ecosystem functioning under climate change is a long-standing problem in global ecology~\cite{reichstein.etal_2014a}; areas of common functioning could be derived directly from SAE-disentangled Earth embeddings, supporting novel science.

Materialising this vision requires several research avenues beyond the scope of this paper.
First, SAE architectures that incorporate geographical principles may retrieve interpretable features more reliably, especially since recent work has identified shortcomings in interpreting SAE features of language models~\cite{chanin.etal_2025, engels.etal_2025, heap.etal_2025}.
Second, auto-interpretability of geospatial features, for instance via Earth observation vision-language models~\cite{kuckreja.etal_2024, zhang.etal_2024a}, could evaluate novel candidate features without labels.
Third, testing SAE features across time~\cite{young.keshav_2026} and on location embeddings such as SatCLIP~\cite{klemmer.etal_2025a, betti.etal_2026} would show whether our conclusions generalise beyond pixel-level self-supervised models.

\ifcameraready
\FloatBarrier
\clearpage

\section*{Acknowledgments}

This work grew out of a discussion with Alexander Shevchenko at the SML lab retreat.
We thank him and the members of the EarthNet team, especially Sebastian Hoffmann, Markus Zehner, Marieke Wesselkamp and Hakam Shams, for their support and for many generous discussions.
The data downloads owe much to Sebastian Hoffmann and Markus Zehner's work on eodata.
We thank Luca Schmidt for discussion and proofreading.
This work has received funding from the European Union under the Horizon Europe project WeatherGenerator (grant agreement 101187947).

This work rests entirely on openly shared data, and we thank the many teams who produce, curate and release them.
The AlphaEarth Foundations Satellite Embedding dataset is produced by Google and Google DeepMind~\cite{brown.etal_2025a}; Tessera~\cite{feng.etal_2025b}, Copernicus-FM~\cite{wang.etal_2025e}, OlmoEarth~\cite{herzog.etal_2025b}, MMEarth~\cite{nedungadi.etal_2025a} and the MajorTOM grid~\cite{francis.czerkawski_2024} were released by their authors.
We thank the producers of the reference products: ESA WorldCover~\cite{zanaga.etal_2022}, ESA CCI Land Cover~\cite{defourny_2019}, Esri Land Cover by Impact Observatory, Microsoft and Esri~\cite{karra.etal_2021}, GLC\_FCS30D~\cite{zhang.etal_2024}, NASA GLanCE~\cite{stanimirova.etal_2023} and MODIS land cover~\cite{sulla-menashe.etal_2019}, ESA WorldCereal~\cite{vantricht.etal_2023}, GEDTM30~\cite{ho.etal_2025}, ISRIC SoilGrids~\cite{poggio.etal_2021}, the Local Climate Zones map~\cite{demuzere.etal_2022}, the Overture Maps Foundation and the OpenStreetMap contributors~\cite{overture_2024}, Meta and the World Resources Institute for CHMv2~\cite{brandt.etal_2026}, the Copernicus DEM~\cite{copernicus_dem_2021}, Global Pasture Watch~\cite{consoli.etal_2025}, and the producers of the CHELSA climate zones, the RESOLVE biomes and the potential natural vegetation map.
Contains modified Copernicus Sentinel data (2024).
\copyright~ESA WorldCover project 2021 / Contains modified Copernicus Sentinel data (2021) processed by ESA WorldCover consortium.
\copyright~ESA Climate Change Initiative - Land Cover led by UCLouvain (2017).
\copyright~OpenStreetMap contributors, Overture Maps Foundation.
Elevation produced using Copernicus WorldDEM-30 \copyright~DLR e.V. 2010-2014 and \copyright~Airbus Defence and Space GmbH 2014-2018 provided under COPERNICUS by the European Union and ESA; all rights reserved.
Qualitative patch panels in Figure~\ref{fig:qualitative} (and the supplementary cross-patch galleries) use the Esri World Imagery basemap.
Source: Esri, Maxar, Earthstar Geographics, and the GIS User Community.
\fi

\section*{References}
\printbibliography[heading=none]

\clearpage

\appendix
\setcounter{figure}{0}
\renewcommand{\thefigure}{S\arabic{figure}}
\setcounter{table}{0}
\renewcommand{\thetable}{S\arabic{table}}

\section*{Supplementary Material}\label{sec:supp}

Appendix~\ref{sec:supp_impl} gives implementation details, including three further Earth embeddings (Appendix~\ref{sec:supp_setup_products}), whose results appear in the reconstruction and concept-alignment appendices.
Appendices~\ref{sec:supp_recon} to~\ref{sec:supp_aoa_results} give additional results for the subsections of §\ref{sec:results}, in the same order: reconstruction (Appendix~\ref{sec:supp_recon}), spatial scale, including the dominant features (Appendix~\ref{sec:supp_spatial}), concept alignment (Appendix~\ref{sec:supp_alignment}), aligned features and sub-concepts (Appendix~\ref{sec:supp_subconcepts}) and area of applicability (Appendix~\ref{sec:supp_aoa_results}).
Each of them begins with the details of the corresponding main-text figure.
Appendix~\ref{sec:broader_impacts} discusses broader impacts.

\section{Implementation Details}\label{sec:supp_impl}

\subsection{Dataset}\label{sec:supp_data}

\begin{figure}[p]
    \centering
    \includegraphics[width=\linewidth,height=0.95\textheight,keepaspectratio]{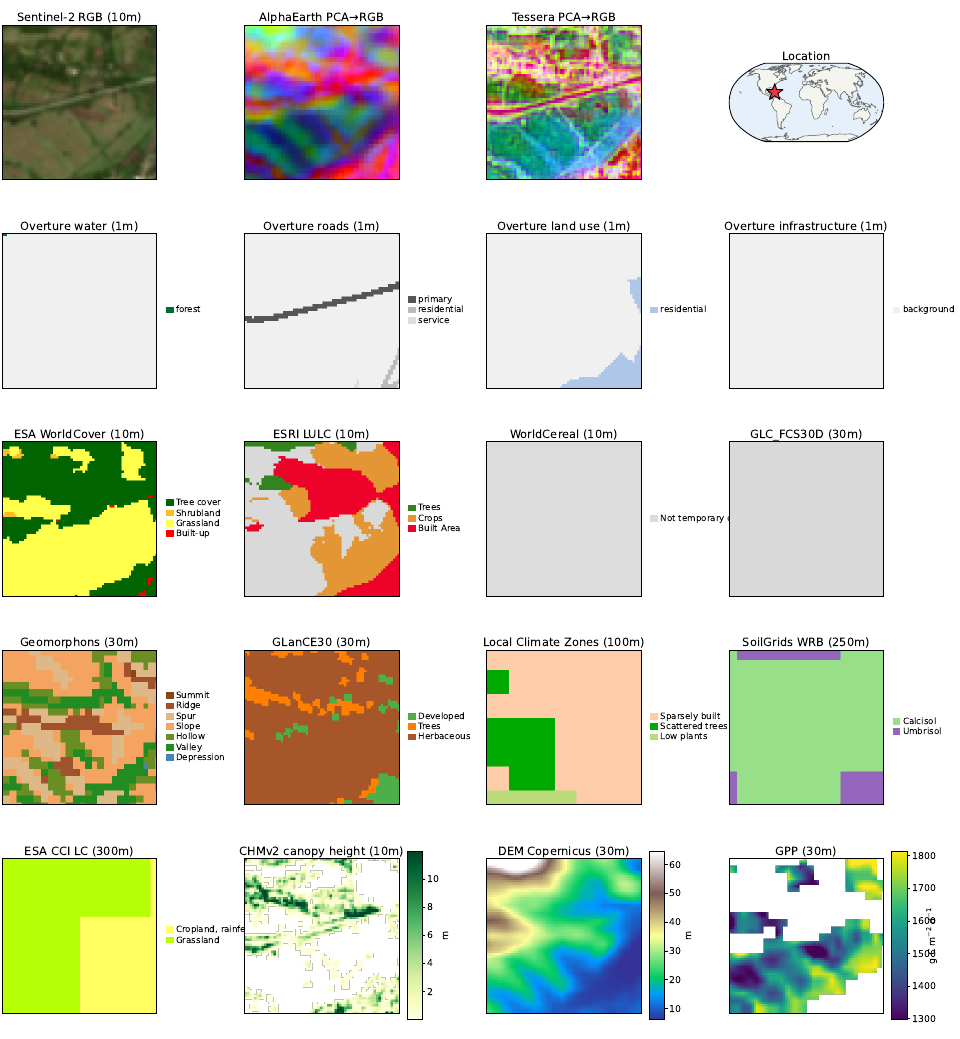}
    \caption{\textbf{Dataset overview on a single representative patch.}
    Top row: Sentinel-2 true-colour image, the first three principal components of the AlphaEarth and Tessera embeddings shown as RGB, and the location of the patch (red star) among all $210{,}000$ sampled patches (grey dots).
    Rows 2--5: the reference products, each with its own legend or colourbar.
    They are 13 of the 14 categorical land-cover, land-use, soil and climate-zone products at $1$\,m to $300$\,m native resolution (the $500$\,m MODIS IGBP map is not shown) and the 3 continuous products (canopy height, elevation, GPP).
    We chose the patch among the test patches with data in all three continuous products, for its land-cover diversity (the Shannon entropy of its ESA WorldCover classes plus the number of its ESRI classes), so that most products show visible structure.}
    \label{fig:supp_dataset_overview}
\end{figure}

We sample $210{,}000$ patches of $64{\times}64$ pixels ($640{\times}640$\,m at $10$\,m resolution) from the about $5.4$ million land cells of the MajorTOM grid~\cite{francis.czerkawski_2024}, in the native Sentinel-2 UTM projection.
The grid cells are spaced $5$\,km apart, which leaves gaps of about $4$\,km between adjacent patches.
For each patch we take the 2024 embeddings of AlphaEarth ($D{=}64$)~\cite{brown.etal_2025a} and Tessera ($D{=}128$)~\cite{feng.etal_2025b}.
We split the $210{,}000$ patches into disjoint sets of $100{,}000$ training, about $10{,}000$ validation and $100{,}000$ test patches ($99{,}922$, $9{,}997$ and $99{,}935$ of them with data), and report all numbers on the test patches.
Our registry holds $14$ categorical and $3$ continuous reference products, from vector layers to $500$\,m land-cover grids, and concept alignment (Appendix~\ref{sec:supp_eval}) uses the categorical ones.
Table~\ref{tab:supp_data_sources} summarises every data source, and Figure~\ref{fig:supp_dataset_overview} shows one patch rendered in every reference product.

\begin{table}[h]
\centering
\footnotesize
\setlength{\tabcolsep}{4pt}
\caption{Data sources used in this paper: two pixel-level Earth embeddings and a registry of 14 categorical and 3 continuous reference products; single-feature alignment is scored against the categorical ones.
Every product is resampled onto the $10$\,m grid of each patch, categorical products by nearest neighbour and continuous ones bilinearly.
The Overture class counts exclude the background class.}
\label{tab:supp_data_sources}
\begin{tabularx}{\linewidth}{@{}llllXl@{}}
\toprule
Name & Type & Resolution & Dim. / classes & Description & Reference \\
\midrule
\multicolumn{6}{@{}l}{\emph{Earth embeddings}} \\
AlphaEarth Foundations          & embedding   & $10$\,m   & $D{=}64$    & Pixel embedding from Sentinel-1/2 and Landsat imagery, trained to also reconstruct GEDI lidar, MODIS land-surface temperature, ERA5 climate, elevation and other sources; depends on the surrounding pixels. & \cite{brown.etal_2025a} \\
Tessera                          & embedding   & $10$\,m   & $D{=}128$   & Pixel embedding from the annual Sentinel-1/2 time series of each single pixel, trained with Barlow Twins; does not look at neighbouring pixels. & \cite{feng.etal_2025b} \\
\midrule
\multicolumn{6}{@{}l}{\emph{Categorical reference products}} \\
Overture water, natural land    & categorical & $1$\,m    & $12$ classes & Water and natural-land vector layer (ocean, lake, river, canal, reservoir, wetland, forest, glacier, sand, bare rock, shrub, grass) from Overture Maps. & \cite{overture_2024} \\
Overture roads                   & categorical & $1$\,m    & $8$ classes  & Road-class vector layer (motorway, primary, secondary, residential, \ldots). & \cite{overture_2024} \\
Overture land use                & categorical & $1$\,m    & $28$ classes & Land-use vector layer (residential, industrial, commercial, farmland, parks, \ldots). & \cite{overture_2024} \\
Overture infrastructure          & categorical & $1$\,m    & $15$ classes & Infrastructure vector layer (airports, ports, power, education, \ldots). & \cite{overture_2024} \\
ESA WorldCover                   & categorical & $10$\,m   & $11$ classes & Global land-cover map for 2021 from Sentinel-1/2 (trees, shrubland, cropland, built-up, water, \ldots). & \cite{zanaga.etal_2022} \\
ESRI Land Cover                  & categorical & $10$\,m   & $9$ classes  & Global Sentinel-2 deep-learning land cover (water, trees, grass, crops, built, bare, snow/ice, \ldots). & \cite{karra.etal_2021} \\
WorldCereal temporary crops      & categorical & $10$\,m   & $2$ classes  & Seasonal binary crop / no-crop mask from the ESA WorldCereal system. & \cite{vantricht.etal_2023} \\
GLC\_FCS30D                      & categorical & $30$\,m   & $35$ classes & Global $30$\,m land cover with a fine classification system, map for $2022$. & \cite{zhang.etal_2024} \\
GEDTM30 geomorphons              & categorical & $30$\,m   & $10$ classes & Pattern-recognition landform classes (peak, ridge, slope, valley, \ldots) from the GEDTM30 DEM. & \cite{ho.etal_2025, jasiewicz.stepinski_2013} \\
GLanCE                           & categorical & $30$\,m   & $7$ classes  & NASA MEaSUREs Global Land Cover Mapping and Estimation, $7$-class scheme. & \cite{stanimirova.etal_2023} \\
Local Climate Zones              & categorical & $100$\,m  & $17$ classes & Stewart--Oke urban-morphology / climate-zone typology, global map for $2018$. & \cite{demuzere.etal_2022} \\
SoilGrids WRB                    & categorical & $250$\,m  & $30$ classes & World Reference Base reference soil groups, predicted globally. & \cite{poggio.etal_2021} \\
ESA CCI Land Cover               & categorical & $300$\,m  & $37$ classes & Climate Change Initiative land cover, map for $2020$. & \cite{defourny_2019} \\
MODIS IGBP land cover            & categorical & $500$\,m  & $17$ classes & MODIS MCD12Q1 land-cover type, IGBP scheme, for $2021$. & \cite{sulla-menashe.etal_2019} \\
\midrule
\multicolumn{6}{@{}l}{\emph{Continuous reference products}} \\
CHMv2 canopy height              & continuous  & $1$\,m    & m            & Global canopy-height map predicted from high-resolution imagery by a vision transformer. & \cite{brandt.etal_2026} \\
Copernicus DEM (GLO-30)          & continuous  & $30$\,m   & m            & Global digital elevation model from Copernicus / Airbus. & \cite{copernicus_dem_2021} \\
Global Pasture Watch GPP         & continuous  & $30$\,m   & g\,C\,m$^{-2}$\,d$^{-1}$ & Bimonthly grassland gross primary productivity from a light-use-efficiency model on Landsat. & \cite{consoli.etal_2025} \\
\bottomrule
\end{tabularx}
\end{table}

\subsection{Concept-group taxonomy}\label{sec:supp_concept_groups}

The reference products overlap strongly: several of them have their own ``water'', ``forest'' or ``built-up'' class, with slightly different definitions and boundaries.
We therefore group the classes of different products that denote the same coarse geographic concept into $20$ hand-curated concept groups (Table~\ref{tab:supp_concept_groups}).
A feature can then align with a concept rather than with one product's version of it, so the evaluation asks which concept a feature represents, not which product it reproduces.

A handful of broad ecosystem-level groups (forest, shrubland, grassland, cropland, built-up, water, wetland, snow / ice, bare, savanna, soil) accumulate many member classes from many products, while finer or single-product groups (e.g.\ moss / lichen, recreation, the five geomorphon groups) contain only a few.
The groups are also not independent, because some ecosystems occur mainly on certain soils or landforms: peat soils (Histosols) co-occur with wetlands, and valleys with riparian forest.
Some alignment off the diagonal of the heatmaps in Figure~\ref{fig:monosemanticity} therefore reflects co-occurrence on the ground rather than polysemantic features.
\begin{table}[h]
\centering
\footnotesize
\setlength{\tabcolsep}{4pt}
\caption{Concept-group taxonomy used in Figure~\ref{fig:monosemanticity}: each concept group aggregates one or more reference-product classes that all denote the same coarse geographic concept.
Multiple classes from the same product are grouped as \texttt{Product: [class$_1$, class$_2$, \ldots]}; long class names are abbreviated.
A feature is \emph{aligned} with a concept group if its folded single-feature AUROC is at least $0.65$ on at least one member class.}
\label{tab:supp_concept_groups}
\begin{tabularx}{\linewidth}{@{}lX@{}}
\toprule
Concept group & Member \texttt{(product, classes)} \\
\midrule
forest & ESA WorldCover: Trees; ESRI: Trees; MODIS IGBP: [Evergreen needleleaf, Evergreen broadleaf, Deciduous needleleaf, Deciduous broadleaf, Mixed]; ESA CCI LC: [Broadleaf evergreen forest, Broadleaf deciduous forest, Broadleaf deciduous closed, Broadleaf deciduous open, Needleleaf evergreen forest, Needleleaf evergreen closed, Needleleaf evergreen open, Needleleaf deciduous forest, Needleleaf deciduous closed, Needleleaf deciduous open, Mixed-leaf forest]; Overture water: forest; Overture land use: forest; LCZ: Dense trees; GLanCE: Trees; GLC\_FCS30D: [Open evergreen broadleaf, Closed evergreen broadleaf, Open deciduous broadleaf, Closed deciduous broadleaf, Open evergreen needleleaf, Closed evergreen needleleaf, Open deciduous needleleaf, Closed deciduous needleleaf, Open mixed-leaf, Closed mixed-leaf] \\
\addlinespace[2pt]
shrubland & ESA WorldCover: Shrubland; ESRI: Scrub; MODIS IGBP: [Closed shrubs, Open shrubs]; ESA CCI LC: [Shrubland, Shrubland, evergreen, Shrubland, deciduous]; Overture water: shrub; LCZ: Bush / scrub; GLanCE: Shrublands; GLC\_FCS30D: [Shrubland, Evergreen shrubland, Deciduous shrubland] \\
\addlinespace[2pt]
grassland & ESA WorldCover: Grassland; ESRI: Grass; MODIS IGBP: Grasslands; ESA CCI LC: Grassland; Overture water: grass; Overture land use: [meadow, recreation\_ground]; LCZ: Low plants; GLanCE: Herbaceous; GLC\_FCS30D: Grassland \\
\addlinespace[2pt]
cropland & ESA WorldCover: Cropland; ESRI: Crops; MODIS IGBP: [Croplands, Cropland / vegetation mosaic]; ESA CCI LC: [Cropland, rainfed, Herbaceous cover, Tree/shrub cover, Cropland, irrigated, Mosaic cropland]; Overture land use: [farmland, orchard, vineyard, allotments, plant\_nursery]; WorldCereal: Temporary crop; GLC\_FCS30D: [Rainfed cropland, Herb. cropland, Tree/shrub cropland, Irrig. cropland] \\
\addlinespace[2pt]
built-up & ESA WorldCover: Built-up; ESRI: Built; MODIS IGBP: Urban; ESA CCI LC: Urban areas; Overture land use: [residential, industrial, commercial, military, construction, retail, education, religious]; LCZ: [Compact high-rise, Compact midrise, Compact low-rise, Open high-rise, Open midrise, Open low-rise, Light low-rise, Large low-rise, Sparsely built, Heavy industry]; GLanCE: Developed; GLC\_FCS30D: Impervious surfaces \\
\addlinespace[2pt]
water & ESA WorldCover: Permanent water; ESRI: Water; MODIS IGBP: Water Bodies; ESA CCI LC: Water bodies; Overture water: [ocean, lake, river, canal, reservoir]; Overture land use: [port, aquaculture]; LCZ: Water; GLanCE: Water; GLC\_FCS30D: Water body \\
\addlinespace[2pt]
wetland & ESA WorldCover: [Herb. wetland, Mangroves]; ESRI: Flooded veg.; MODIS IGBP: Wetland; ESA CCI LC: [Tree, flooded fresh/brackish, Tree, flooded saline, Shrub/herb, flooded]; Overture water: wetland; GLC\_FCS30D: [Swamp, Marsh, Flooded flat, Saline, Mangrove, Salt marsh, Tidal flat] \\
\addlinespace[2pt]
snow / ice & ESA WorldCover: Snow/ice; ESRI: Snow/Ice; MODIS IGBP: Snow/ice; ESA CCI LC: Snow/ice; Overture water: glacier; Overture land use: winter\_sports; GLanCE: Ice / snow; GLC\_FCS30D: Permanent ice and snow \\
\addlinespace[2pt]
bare & ESA WorldCover: Bare; ESRI: Bare Ground; MODIS IGBP: Barren; ESA CCI LC: [Sparse vegetation, Sparse tree, Sparse shrub, Sparse herbaceous, Bare, Consol. bare, Unconsol. bare]; Overture water: [sand, bare\_rock]; Overture land use: [quarry, landfill]; LCZ: [Bare rock / paved, Bare soil / sand]; GLanCE: Barren; GLC\_FCS30D: [Sparse veg., Sparse shrub, Sparse herb., Bare, Consol. bare, Unconsol. bare] \\
\addlinespace[2pt]
savanna & MODIS IGBP: [Woody savanna, Savannas]; ESA CCI LC: [Mosaic tree/shrub, Mosaic herb, Mosaic veg/crop]; LCZ: Scattered trees \\
\addlinespace[2pt]
moss / lichen & ESA WorldCover: Moss / lichen; ESA CCI LC: Lichens / mosses; GLC\_FCS30D: Lichens / mosses \\
\addlinespace[2pt]
road & Overture roads: [motorway, trunk, primary, secondary, tertiary, residential, service, railway]; Overture land use: railway \\
\addlinespace[2pt]
infrastructure & Overture infra.: [runway, taxiway, apron, bridge, dam, pier, substation, tower, solar\_panel\_area, water\_tower, power\_line, pipeline, communication\_tower, waste\_disposal, other] \\
\addlinespace[2pt]
recreation & Overture land use: [park, stadium, golf\_course, zoo] \\
\addlinespace[2pt]
flat & Geomorphons: Flat \\
\addlinespace[2pt]
summit & Geomorphons: Summit \\
\addlinespace[2pt]
ridge & Geomorphons: [Ridge, Spur] \\
\addlinespace[2pt]
slope & Geomorphons: [Shoulder, Slope] \\
\addlinespace[2pt]
valley & Geomorphons: [Hollow, Footslope, Valley, Depression] \\
\addlinespace[2pt]
soil & SoilGrids WRB: [Acrisol, Albeluvisol, Alisol, Andosol, Arenosol, Calcisol, Cambisol, Chernozem, Cryosol, Durisol, Ferralsol, Fluvisol, Gleysol, Gypsisol, Histosol, Kastanozem, Leptosol, Lixisol, Luvisol, Nitisol, Phaeozem, Planosol, Plinthosol, Podzol, Regosol, Solonchak, Solonetz, Stagnosol, Umbrisol, Vertisol] \\
\addlinespace[2pt]
\bottomrule
\end{tabularx}
\end{table}

\subsection{Evaluation Protocol}\label{sec:supp_eval}

This subsection gives the exact formula and the evaluation sample of each of the three criteria of §\ref{sec:interpretability}.

\textit{Reconstruction at matched sparsity.} For an embedding $\mathbf{e}\in\mathbb{R}^D$ with SAE reconstruction $\hat{\mathbf{e}}$, and with $\bar{\mathbf{e}}$ the mean test embedding, we report the mean-squared error per embedding dimension, the fraction of variance explained (with the variance summed over dimensions) and the cosine similarity:
\begin{equation}
\mathrm{MSE} = \frac{1}{D}\,\mathbb{E}\,\|\mathbf{e}-\hat{\mathbf{e}}\|_2^2,\qquad
\mathrm{FVE} = 1 - \frac{\mathbb{E}\|\mathbf{e}-\hat{\mathbf{e}}\|_2^2}{\mathbb{E}\|\mathbf{e}-\bar{\mathbf{e}}\|_2^2},\qquad
\mathrm{cos} = \mathbb{E}\,\frac{\langle \mathbf{e},\hat{\mathbf{e}}\rangle}{\|\mathbf{e}\|\,\|\hat{\mathbf{e}}\|}.
\label{eq:supp_recon}
\end{equation}
Here $\mathbb{E}$ is the mean over a fixed sample of $199{,}870$ test pixels, two per test patch; for reconstruction we leave out the $1{,}335$ AlphaEarth pixels that carry no data (a constant fill value).

We compare these scores at matched sparsity $L_0=\mathbb{E}\,\|\mathbf{z}\|_0$, the mean number of non-zero features per pixel over the same pixel sample.

\textit{Concept alignment.} For every feature $j$ and every class $c$ of a categorical reference product, we use the activation $z_j$ alone as a score that separates pixels of class $c$ from other pixels, on a balanced sample of up to $2{,}500$ pixels of class $c$ and as many other pixels, both drawn from the same $199{,}870$ test pixels.
Its AUROC is $A_{jc}=\Pr(z^{+}>z^{-})$, the probability that the activation $z^{+}$ on a random pixel of class $c$ exceeds the activation $z^{-}$ on a random other pixel; it equals a normalised Mann--Whitney $U$ statistic.
We use its folded form
\begin{equation}
    \mathrm{AUROC}^{\pm}(z_j; c) = \max\big(A_{jc},\, 1-A_{jc}\big),
    \label{eq:folded_auroc}
\end{equation}
so that a feature that separates $c$ by staying low on it also counts as aligned.
For a concept group $g$ we write $\mathrm{AUROC}^{\pm}(z_j;g)=\max_{c\in g}\mathrm{AUROC}^{\pm}(z_j;c)$, and feature $j$ is \emph{aligned} with $g$ if $\mathrm{AUROC}^{\pm}(z_j;g)\geq 0.65$.
The alignment counts use only \emph{live} features, whose AUROC exceeds $0.51$ for at least one class.
The coverage count UCov instead uses product classes at $0.60$ (Appendix~\ref{sec:supp_alignment_scores}, Eq.~\ref{eq:ucov_emf}).

\textit{Spatial scale.} We estimate the semivariance of Eq.~\ref{eq:variogram} on a nested sample of test patches.
We draw $400$ anchor patches at random and add, for each anchor, its $1$st, $2$nd, $3$rd, $5$th, $8$th, $12$th, $18$th and $27$th nearest test patch ($3{,}543$ patches in total).
Short and long separations thus come from the same random places.
Features are computed on all $64{\times}64$ pixels of every patch, after the input normalisation of each SAE (Appendix~\ref{sec:supp_training}).
The baseline decompositions are fitted with all $D$ components on $100{,}000$ pixels of this sample.
Within a patch, $\gamma_j(h)$ is half the mean squared difference of feature $j$ over all horizontal and vertical pixel pairs at separation $h\in\{1,2,3,4,6,8,12,16,24,32,48,63\}$ pixels, that is $10$ to $630$\,m.
Between two patches $p$ and $q$, in which feature $j$ has pixel means $m_p, m_q$ and pixel variances $v_p, v_q$, the semivariance over all $4{,}096{\times}4{,}096$ pixel pairs is exactly $\tfrac12[(m_p-m_q)^2+v_p+v_q]$, so it needs no pairwise computation.
Patch pairs are binned by the great-circle distance of their centres into twelve log-spaced bins from $4$ to $20{,}100$\,km, and a bin is shown only if it contains at least $300$ patch pairs (the first bin shown is at $13$\,km).
The variance $\sigma_j^2$ is the variance of feature $j$ over all pixels of the sample, that is, the mean within-patch variance plus the variance of the patch means; features with $\sigma_j^2<10^{-10}$ are dropped.
Because the normalised semivariance need not rise monotonically, the correlation length is taken at its first crossing of $1-1/e$, interpolated in log distance between the two sampled separations on either side.
If these two separations lie on either side of the band between $640$\,m and $13$\,km, which no pixel pair samples, the correlation length is reported as falling in this band.

\textit{Patch-mean Moran's $I$.} As a second, standard measure of spatial autocorrelation, we use Moran's $I$ on a fixed sample of $N{=}1{,}500$ test patches.
Let $\bar{z}_{i,j}$ be the mean activation of feature $j$ over all $64{\times}64$ pixels of patch $i$, and $\bar{z}_j$ its mean over all $N$ patches.
Raw embedding dimensions are averaged in the same way, and each baseline decomposition is fitted to these $1{,}500$ patch-mean embeddings and applied to them.
The weight $w_{ii'}$ is $1/8$ if patch $i'$ is one of the $8$ patches whose centres lie nearest to that of patch $i$, and $0$ otherwise.
The patch-mean Moran's $I$ of feature $j$ is then
\begin{equation}
I_j \;=\; \frac{\sum_{i,i'}w_{ii'}\,(\bar{z}_{i,j}-\bar{z}_j)\,(\bar{z}_{i',j}-\bar{z}_j)}{\sum_i (\bar{z}_{i,j}-\bar{z}_j)^2},
\label{eq:supp_moran}
\end{equation}
which is close to $1$ if neighbouring patches have similar means and close to $0$ if they are unrelated.
We report its median over all features whose patch mean varies over the $1{,}500$ patches.
Moran's $I$ thus covers every feature whose patch mean varies, whereas the alignment counts cover only live features, so the two analyses use slightly different sets of features.

\paragraph{Sample sizes.} The test patches hold about $4{\times}10^8$ pixels, and scoring every model on all of them would cost more than training it.
Each criterion therefore uses one of the fixed samples above, and all models are scored on the same samples, so that differences between them reflect the models and not sampling noise.

\subsection{Dominant features}\label{sec:supp_energy}

Beyond the three criteria, we ask which features dominate a sparse code, that is, which carry most of the squared activations.
The \emph{energy share} of feature $j$ is the mean of $z_j^2$ divided by the mean of $\|\mathbf{z}\|_2^2$, both over $200{,}000$ training pixels, two from each training patch.
We compare the atoms of the features of largest energy share with the leading principal components of the embedding by their absolute cosine similarity, absolute because the sign of a principal component is arbitrary (Appendix~\ref{sec:supp_dominant}).

\subsection{Training Details}\label{sec:supp_training}
\paragraph{SAE architecture.} Every SAE maps a pixel's embedding to a sparse code and back in three steps.
\begin{enumerate}[leftmargin=1.4em]
\item \emph{Encoder.} A single linear map $\mathbf{W}_\text{enc}\in\mathbb{R}^{M\times D}$ with bias $\mathbf{b}_\text{enc}$ computes $M$ pre-activations, one per atom.
The decoder bias $\mathbf{b}$ is also subtracted from the embedding $\mathbf{e}$ before encoding~\cite{bricken2023monosemanticity}, so that the encoder computes $\mathbf{W}_\text{enc}(\mathbf{e}-\mathbf{b})+\mathbf{b}_\text{enc}$.
This only reparametrises Eq.~\ref{eq:sae}, because $\mathbf{W}_\text{enc}\mathbf{b}$ can be absorbed into $\mathbf{b}_\text{enc}$.
The decoder starts from the default initialisation of a linear layer in PyTorch, with each atom then scaled to unit norm, and the encoder from $\mathbf{W}_\text{enc}=\mathbf{W}_\text{dec}^\top$ and $\mathbf{b}_\text{enc}=\mathbf{0}$; both are trained freely afterwards.
AlphaEarth embeddings, which have unit norm, are used as released.
Tessera embeddings are standardised per dimension; ReLU+$\ell_1$ takes them as input at this scale (mean norm about $\sqrt{D}$), and the BatchTopK variants after rescaling to about unit norm.
\item \emph{Sparsifier.} The sparsifier $\mathcal{S}$ decides which pre-activations survive.
BatchTopK and Matryoshka keep the $B{\cdot}k$ largest across the $B$ pixels of a training batch, so $k$ sets the average number of active features per pixel; the selection acts directly on the pre-activations, without a preceding ReLU.
The threshold that replaces this selection at test time is a running average (decay $0.999$) of the smallest kept activation of each training batch.
ReLU+$\ell_1$ instead passes all of them through a ReLU and adds the sum of the activations, times an $\ell_1$ coefficient, to the loss.
This coefficient sets $L_0$ only indirectly, so it is adapted during training until $L_0$ matches $k$.
\item \emph{Decoder.} The surviving activations weight the columns of a single linear map $\mathbf{W}_\text{dec}\in\mathbb{R}^{D\times M}$, the atoms, which are renormalised to unit norm before every training step, so that the size of each term of the reconstruction is set by its activation and not by the length of its atom.
A learned bias $\mathbf{b}$ is added to give $\hat{\mathbf{e}}$.
\end{enumerate}
The loss is the squared reconstruction error $\mathcal{L}_\text{rec}$ of Eq.~\ref{eq:sae}.
The BatchTopK and Matryoshka variants add an auxiliary loss $\mathcal{L}_\text{aux}$ with weight $\beta{=}1/32$, in which the $k_\text{aux}{=}32$ largest pre-activations of dead features, those that have not been active for $25$ batches, reconstruct the residual $\mathbf{e}-\hat{\mathbf{e}}$.
Matryoshka adds a second reconstruction term in which only the first $n_s = \mathrm{round}(0.2M)$ atoms ($205$ at $M{=}1024$), the \emph{prefix}, reconstruct $\mathbf{e}$, with the same weight as the full reconstruction:
\begin{equation}
    \mathcal{L}_\text{Matryoshka} = \mathcal{L}_\text{rec} + \big\|\mathbf{e} - (\mathbf{W}_\text{dec}^{[:n_s]}\mathbf{z}^{[:n_s]} + \mathbf{b})\big\|_2^2 + \beta\,\mathcal{L}_\text{aux}.
    \label{eq:matryoshka}
\end{equation}
This term pushes coarse, frequently used concepts into the prefix.
The frozen-decoder baseline (§\ref{sec:interpretability}) is a BatchTopK SAE whose decoder columns are drawn once as random unit-norm vectors and never updated, with the decoder bias fixed at zero; its encoder and test-time threshold are trained and estimated as for the other SAEs.

\paragraph{Optimisation.} We optimise the SAE objective with Adam ($\beta_1{=}0$, $\beta_2{=}0.999$, no weight decay) at a peak learning rate of $10^{-3}$, with a $1{,}000$-step linear warmup and cosine annealing, and clip gradients to a global norm of $1$.
Every model trains for $50$ epochs of $1{,}000$ steps each (an epoch here is a fixed number of steps, not a pass over the data), and we keep the checkpoint with the best validation reconstruction.

\paragraph{Batch composition.} A training step draws $512$ anchor patches from the training patches held in memory, $8$ anchor pixels in each with two nearby pixels per anchor, and $128$ uniform pixels from every loaded patch: about $7.7{\times}10^{4}$ pixels, which is the batch over which the BatchTopK rule selects its $B{\cdot}k$ largest pre-activations.
The same batch composition is used for every SAE.

\paragraph{Compute costs.} The training set is held in memory, and every run uses a single GPU (NVIDIA A100 or A40).
The sweep of Figure~\ref{fig:pareto} aggregates $90$ SAE training runs (the three variants at $k\in\{8, 16, 32\}$ for $M{=}1024$ and at $M\in\{512, 2048\}$ for $k{=}16$, three seeds each, on AlphaEarth and Tessera) and $18$ frozen-decoder runs.
By the logged training time, a run takes a median of $22$\,min on AlphaEarth and $58$\,min on Tessera (range $15$ to $73$\,min), and the $108$ runs total $73$ GPU-hours.
Evaluation and preliminary experiments are not included in this total; their cost was not logged.

\subsection{Baselines}\label{sec:supp_baselines}

The baselines test whether a change of basis of the embedding already gives interpretable axes, or whether an overcomplete, sparse dictionary is needed.
PCA, ICA and NMF are the decompositions a practitioner would try first, and each embodies a different notion of a good axis.
Each has $D$ components, one per embedding dimension.

\paragraph{PCA.} Principal component analysis finds orthogonal directions of decreasing variance.
We centre the embedding and do not whiten it.
PCA is the natural first look at any embedding, and it tests whether the directions of largest variance are also the directions of single concepts.

\paragraph{ICA.} Independent component analysis finds directions along which the projected data are maximally non-Gaussian, and hence as statistically independent as possible.
We use FastICA with unit-variance whitening and at most $500$ iterations.
If concepts were independent sources mixed linearly into the embedding, ICA would recover them, so among the decompositions with $D$ components it is the strongest competitor for monosemanticity.

\paragraph{NMF.} Non-negative matrix factorisation writes the data as a product of non-negative atoms and non-negative codes, the same constraint an SAE places on its activations.
Because Earth embeddings are signed, we first subtract from every dimension its minimum over the sample, and we run at most $400$ iterations.
NMF therefore tests whether non-negativity alone, without overcompleteness and sparsity, yields parts-based features.

\paragraph{Fitting.} For the per-feature analyses, each decomposition is fitted on the sample of the analysis itself (Appendix~\ref{sec:supp_eval}), ICA and NMF on a $30{,}000$-pixel subsample of it, and its component scores are evaluated as features in the same way as SAE activations.
For the reconstruction of Figure~\ref{fig:pareto}, the bases are fitted on $100{,}000$ training pixels, one from each training patch, and scored on the test sample of the SAEs ($4{,}000$ test patches with $256$ pixels each, without patches of constant fill or with no-data pixels) with the same FVE computation.
Each baseline keeps $k$ components, so that its $L_0$ matches that of the SAEs at the same $k$.
PCA keeps the first $k$ principal components, and ICA and NMF are fitted with $k$ components.
Because FastICA first whitens with PCA, the $k$ ICA components span the same subspace as the first $k$ principal components, and their FVE agrees with that of PCA to within $10^{-3}$.
For the raw embedding, we keep the $k$ dimensions of largest variance on the training pixels and set the others to their training mean.
NMF encodes a pixel by the least-squares coefficients of its shifted embedding, each dimension minus its training minimum, on the NMF atoms.

\subsection{Area-of-Applicability Computation}\label{sec:supp_aoa}

The area-of-applicability (AoA) recipe of §\ref{sec:downstream} ranks held-out pixels by their mean Euclidean distance to their five nearest training pixels in a representation space.
In the raw space a pixel is represented by its embedding, and in the SAE space by its sparse code $\mathbf{z}$.
AlphaEarth embeddings are used as released, since they have unit norm, and Tessera embeddings are standardised per dimension, as for the SAEs (Appendix~\ref{sec:supp_training}).
SAE codes are not rescaled, because an activation carries its magnitude by construction, the atoms having unit norm.
Per-feature rescaling changes which features dominate the distance, and Appendix~\ref{sec:supp_aoa_scaling} reports its effect.
We report two variants of the distance, unweighted and ridge-weighted.

\paragraph{Ridge-weighted distance.} In each fold, a ridge regressor, the \emph{head}, predicts the target from the representation of the training pixels.
The ridge-weighted distance follows the weighting of Meyer and Pebesma~\cite{meyer.pebesma_2021}, but replaces their random-forest variable importance by the absolute coefficient $\lvert\hat{\beta}_j\rvert$ of feature $j$ in this head, relative to the mean over features, $w_j = \lvert\hat{\beta}_j\rvert / \overline{\lvert\hat{\beta}\rvert}$.
The distance of a held-out pixel with representation $\mathbf{x}$ is
\begin{equation}
    d_{w}(\mathbf{x}) \;=\; \frac{1}{5}\!\!\sum_{i \in \mathcal{N}_5(\mathbf{x})}\!\!\sqrt{\sum_j w_j\,(x_j - x_{ij})^2},
\end{equation}
where $\mathbf{x}_i$ is the representation of training pixel $i$ and $\mathcal{N}_5(\mathbf{x})$ the set of the five training pixels nearest to $\mathbf{x}$ under this weighting.
The unweighted distance is the case $w_j=1$, and the weighted distance equals the unweighted one after each feature is scaled by $\sqrt{w_j}$.
We divide the distance by the median of the same quantity over the training pixels, each excluded from its own neighbours; this does not change any rank statistic.
The score of a distance is the Spearman rank correlation $\rho$ between it and the absolute prediction error on the held-out pixels.

\paragraph{Maps of the area of applicability.} Where we report a share of pixels inside or outside the area of applicability, we use the dissimilarity index and threshold of Meyer and Pebesma~\cite{meyer.pebesma_2021}.
The index is the distance to the nearest training pixel divided by the mean distance between training pixels.
The threshold is the upper whisker (third quartile plus $1.5$ interquartile ranges) of the index of the training pixels, where the index of each training pixel is taken against the training pixels outside its own cross-validation fold, with ten folds of $10^\circ$ longitude--latitude blocks.

\paragraph{Sites, folds and checkpoints.} We draw one pixel, a \emph{site}, from each test patch, so that sites lie at least about $4$\,km apart: $98{,}147$ sites with a finite canopy height (CHMv2~\cite{brandt.etal_2026}), elevation (Copernicus DEM~\cite{copernicus_dem_2021}) and MODIS IGBP class.
Four split scenarios partition them: train on one continent and predict the rest; leave one continent out; train on one plant functional type (PFT, the dominant IGBP class of the patch) and predict the rest; leave one PFT out.
Continents are longitude--latitude boxes.
Six continents and seven PFTs give $26$ folds, and the two targets give $52$ combinations of fold and target.
In each fold we fit the head with regularisation strength $\alpha{=}1$ on up to $12{,}000$ training sites, which also serve as the training pixels of the nearest-neighbour search, and we score up to $8{,}000$ held-out sites.
We score four checkpoints at $M{=}1024$, $k{=}16$ on AlphaEarth, BatchTopK at three seeds and the Matryoshka checkpoint of Figure~\ref{fig:aoa}, so each statistic is a distribution over $208$ \emph{cells}, one per combination of fold, target and checkpoint.

\paragraph{Graded hold-outs.} The result of §\ref{sec:results_aoa} uses a second set of folds, graded by how far the held-out sites lie from the training data.
Each fold holds out the $1{,}000$ sites nearest a centre and trains on the $12{,}000$ nearest sites that lie at least a gap beyond them.
In the distance design, $8$ centres at least $2{,}000$\,km apart each get gaps of $0$, $250$, $500$, $1{,}000$, $2{,}000$ and $4{,}000$\,km of great-circle distance.
In the climate design, $8$ other centres each get gaps of $0$, $0.25$, $0.5$, $1$ and $2$ standard deviations in a climate distance, the Euclidean distance between standardised CHELSA annual mean temperature, temperature seasonality, log annual precipitation and precipitation seasonality.
This gives $8\cdot6+8\cdot5=88$ folds per target.
The folds of one centre share their held-out sites, so they are not independent, and intervals from a bootstrap over folds are optimistic.
One ridge head is fitted on the raw embedding, and every distance is scored by how well it ranks the absolute errors of this head, so that the comparison does not depend on how well each representation predicts.
The ridge weights of a distance come from a head fitted in its own representation space.
Figure~\ref{fig:aoa_main} reports the gain in $\rho$ of a sparse code over the raw distance, for a BatchTopK SAE ($M{=}1024$, $k{=}16$, three seeds) and for an untrained random sparse code of the same dictionary size and sparsity (Appendix~\ref{sec:supp_aoa_graded}).
This random code projects the centred embedding onto $M$ random Gaussian unit-norm directions and keeps the $k$ largest positive projections of each pixel.

\subsection{Further Earth embeddings}\label{sec:supp_setup_products}

To test whether our findings transfer beyond AlphaEarth and Tessera, we train BatchTopK SAEs on three further Earth embeddings, which differ from both in width, in the ground area that one vector covers and in how they are produced.

\paragraph{Copernicus-FM}~\cite{wang.etal_2025e} releases a global embedding map with one $D{=}768$ vector per $0.25^\circ$ cell (about $28$\,km), averaged over all Sentinel-1, -2, -3, -5P and Copernicus DEM observations of the cell.
We take every cell that has an embedding and contains at least one patch of the MajorTOM patch collection from which our sample is drawn, split them at random into $169{,}972$ training, $12{,}181$ validation and $60{,}696$ test cells, and evaluate on $4{,}000$ of the test cells.
Averaging moves every cell toward the global mean: the mean cosine similarity between a cell's embedding and the global mean embedding is $0.98$, so a model that outputs only the mean already reaches a reconstruction cosine of $0.98$.
We therefore centre the embeddings on their training mean and divide them by one scale shared by all dimensions, which leaves the FVE unchanged.

\paragraph{OlmoEarth and MMEarth.}
OlmoEarth~\cite{herzog.etal_2025b} (v1-Base, $D{=}768$) and MMEarth~\cite{nedungadi.etal_2025a} (ConvNeXt~V2 atto, $D{=}320$) release models rather than embeddings.
We run them on one Sentinel-2 L2A scene of 2024 per patch, with less than $10\%$ cloud cover and closest to the local peak of the growing season, over the central $64{\times}64$ pixels of $20{,}600$ of our patches ($9{,}811$ for training).
At a stride of $8$ pixels, each patch yields an $8{\times}8$ grid of $80$\,m tokens, and each token is one sample, standardised per dimension with training statistics.
We evaluate on all $64$ tokens of $4{,}000$ test patches.

\paragraph{Training.}
The SAEs are BatchTopK SAEs, trained as in Appendix~\ref{sec:supp_training}, with three seeds per setting unless stated otherwise.
On a wider embedding, the same $k$ is a much smaller share of the dimensions: $k{=}16$ is $25\%$ of the dimensions of AlphaEarth but $2.1\%$ of those of OlmoEarth.
We therefore match the number of active features per dimension, $k/D{=}0.25$, which gives $M{=}1024$ and $k{=}192$ for Copernicus-FM, $M{=}4096$ and $k{=}192$ for OlmoEarth and $M{=}4096$ and $k{=}80$ for MMEarth.
Each setting has a frozen-decoder baseline of the same $M$ and $k$.
For the concept alignment of OlmoEarth and MMEarth, each $80$\,m token carries the majority class of the $8{\times}8$ pixels it covers in every reference product.
We do not score the alignment of Copernicus-FM: at $28$\,km the majority class of a cell covers on average only $75\%$ of its pixels for ESA WorldCover and $48\%$ for geomorphons.

\paragraph{Embeddings that an SAE does not fit.}
An SAE assumes a continuous embedding space in which more concepts than dimensions are superposed, that is, stored as overlapping directions.
Some embeddings do not meet this premise.
The ESD embedding database~\cite{chen.etal_2026} ($30$\,m, multi-decadal, $D{=}12$) stores a quantised codebook index in each of its $12$ channels.
Each channel holds one monthly latent step, chosen to align with the seasonal cycle.
Its channels therefore already have a designed meaning, and its values are discrete.
Neither superposition nor continuity holds for ESD, so decomposing it would need SAEs built for quantised representations, which is beyond this paper.

\subsection{Overview figure}\label{sec:supp_fig1}

\paragraph{Details of Figure~\ref{fig:overview}.} Panel c shows the Matryoshka features f437 (central Australia) and f635 (Tibetan Plateau); the colour is the mean, over the test patches in a $2^\circ{\times}2^\circ$ cell, of the share of the $64{\times}64$ pixels of a patch on which the feature is active.
Neither feature appears in the other figures.

\FloatBarrier
\clearpage
\section{Reconstruction}\label{sec:supp_recon}

This appendix supports §\ref{sec:results_table}.

\paragraph{Details of Figure~\ref{fig:pareto}.} The FVE axis is logarithmic in $1-\mathrm{FVE}$, and the horizontal axis is $L_0$, the mean number of active features per pixel.
Colour encodes the method, marker shape the SAE variant and marker size the dictionary size $M$; lines join $k\in\{8,16,32\}$ at $M{=}1024$, and the three dictionary sizes at $k{=}16$ are offset horizontally for legibility.
The frozen-decoder baseline is the BatchTopK SAE with its decoder frozen at the random initialisation, seed $0$ (the FVE of the three seeds differs by at most $0.012$).
The baselines, each with $k$ components, are those of Appendix~\ref{sec:supp_baselines}, fitted on training pixels and scored on the test pixels of the SAEs; ICA has the same FVE as PCA.
SAE points are means of three seeds (standard deviation at most $0.007$), and the baselines are single fits.
At $k{=}32$ on AlphaEarth, $75\%$ of the Matryoshka and $15\%$ of the BatchTopK atoms are never active, so these SAEs use fewer than $M$ features.

\subsection{Reconstruction at matched sparsity}\label{sec:supp_table}
Table~\ref{tab:eight_criteria} reports the reconstruction metrics at $L_0\approx16$ for the SAEs and all baselines.
On both embeddings, every SAE variant reconstructs better than every baseline, and BatchTopK best.
The frozen-decoder baseline explains more of AlphaEarth than the first $16$ principal components ($0.775$ against $0.696$) and about as much of Tessera ($0.534$ against $0.519$), as in Figure~\ref{fig:pareto}.
ICA spans the same subspace as PCA and therefore scores identically.

\begin{table}[h]
\centering
\footnotesize
\caption{Reconstruction at $L_0\approx16$ on AlphaEarth and Tessera: the three SAE variants ($M{=}1024$, $k{=}16$), the frozen-decoder baseline of the same size and sparsity, and the baselines with $16$ components (Appendix~\ref{sec:supp_baselines}), on the test pixels of Appendix~\ref{sec:supp_eval}.
Best value per embedding in bold.}
\label{tab:eight_criteria}
\begin{tabular}{lcccc}
\toprule
 & MSE ($\downarrow$) & FVE ($\uparrow$) & cos ($\uparrow$) & $L_0$ \\
\midrule
\multicolumn{5}{l}{\emph{AlphaEarth ($D{=}64$)}}\\
Raw ($16$ dims)        & $9.44{\cdot}10^{-3}$ & $0.325$ & $0.623$ & $16$ \\
PCA-$16$               & $4.25{\cdot}10^{-3}$ & $0.696$ & $0.845$ & $16$ \\
ICA-$16$               & $4.25{\cdot}10^{-3}$ & $0.696$ & $0.845$ & $16$ \\
NMF-$16$               & $4.42{\cdot}10^{-3}$ & $0.684$ & $0.838$ & $16$ \\
Frozen decoder         & $3.14{\cdot}10^{-3}$ & $0.775$ & $0.894$ & $16.3$ \\
\addlinespace[2pt]
ReLU+$\ell_1$          & $9.19{\cdot}10^{-4}$ & $0.934$ & $0.975$ & $16.3$ \\
BatchTopK              & $\mathbf{3.72{\cdot}10^{-4}}$ & $\mathbf{0.973}$ & $\mathbf{0.988}$ & $16.1$ \\
Matryoshka             & $4.23{\cdot}10^{-4}$ & $0.970$ & $0.987$ & $16.0$ \\
\midrule
\multicolumn{5}{l}{\emph{Tessera ($D{=}128$)}}\\
Raw ($16$ dims)        & $7.17{\cdot}10^{-3}$ & $0.124$ & $0.348$ & $16$ \\
PCA-$16$               & $3.93{\cdot}10^{-3}$ & $0.519$ & $0.705$ & $16$ \\
ICA-$16$               & $3.93{\cdot}10^{-3}$ & $0.519$ & $0.705$ & $16$ \\
NMF-$16$               & $4.06{\cdot}10^{-3}$ & $0.503$ & $0.694$ & $16$ \\
Frozen decoder         & $3.81{\cdot}10^{-3}$ & $0.534$ & $0.713$ & $16.3$ \\
\addlinespace[2pt]
ReLU+$\ell_1$          & $1.40{\cdot}10^{-3}$ & $0.828$ & $0.918$ & $16.5$ \\
BatchTopK              & $\mathbf{6.65{\cdot}10^{-4}}$ & $\mathbf{0.919}$ & $\mathbf{0.957}$ & $16.3$ \\
Matryoshka             & $7.04{\cdot}10^{-4}$ & $0.914$ & $0.954$ & $16.0$ \\
\bottomrule
\end{tabular}
\end{table}

\subsection{Sparse coding in a fixed basis}\label{sec:supp_fixed_basis}

Choosing the components per pixel is part of what an SAE does.
Keeping each pixel's $k$ largest coefficients out of all $D$ in the PCA basis, so that the kept components differ between pixels, reaches an FVE of $0.75$, $0.88$, $0.98$ on AlphaEarth and $0.56$, $0.72$, $0.86$ on Tessera at $k{=}8$, $16$, $32$.
In the raw basis the same rule reaches $0.48$, $0.71$, $0.92$ and $0.31$, $0.49$, $0.71$.
At $k{=}16$, the PCA-basis rule closes 67\% (AlphaEarth) and 50\% (Tessera) of the FVE gap between the first $16$ principal components and BatchTopK.
This indicates how much of the gain of the SAEs over rank-$k$ PCA comes from choosing the components per pixel rather than from the learned overcomplete dictionary.

\FloatBarrier
\subsection{Reconstruction quality on raw embedding dimensions}\label{sec:supp_recon_quality}

\begin{figure}[ht]
    \centering
    \includegraphics[width=0.6\linewidth]{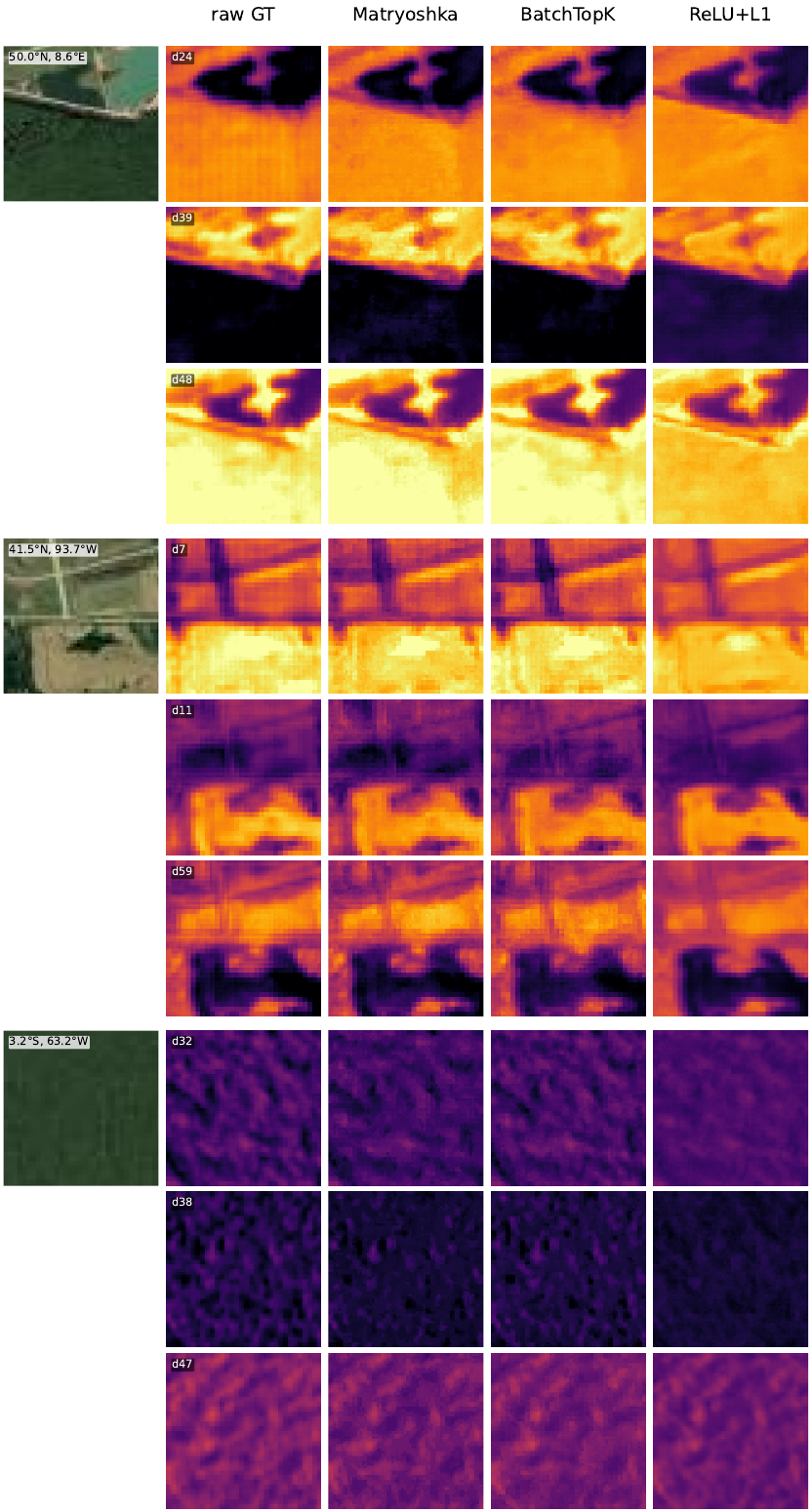}
    \caption{Per-dimension reconstruction quality of the three SAE variants.
    Three AlphaEarth test patches without missing ESA WorldCover pixels, the nearest such patches to fixed locations near Frankfurt, in the US Midwest and in the Amazon, one block each.
    A block shows the Esri imagery tile (top left) and three active raw embedding dimensions of the patch, one per row, with the dimension index in the first column.
    Each row shows the original dimension (GT) and its reconstruction by Matryoshka, BatchTopK and ReLU+$\ell_1$.
    All panels share one colour scale, so a colour mismatch with the GT column is a reconstruction error.}
    \label{fig:supp_recon_quality}
\end{figure}

Figure~\ref{fig:supp_recon_quality} shows per-pixel reconstruction quality on three AlphaEarth patches (a lake edge, farmland and forest), with three of the most active raw embedding dimensions per patch and their SAE reconstructions at $M{=}1024$, $k{=}16$.
All maps share one colour scale, so where a reconstruction is accurate, its colours match those of the original dimension in the column GT (ground truth).
BatchTopK and Matryoshka reproduce both smooth transitions and sharp class boundaries, such as the water edge in patch 1.
ReLU+$\ell_1$ reproduces the same structure at a slightly lower amplitude, consistent with its lower FVE.

\FloatBarrier
\subsection{Further Earth embeddings}\label{sec:supp_recon_products}

Table~\ref{tab:supp_products_recon} compares the reconstruction of BatchTopK SAEs on Copernicus-FM, OlmoEarth and MMEarth (Appendix~\ref{sec:supp_setup_products}) with the frozen-decoder baseline and with PCA and the raw embedding at the same $L_0$, as in Figure~\ref{fig:pareto}.
At both sparsity levels and on all three embeddings, the SAE reconstructs better than every baseline.
On all three, the frozen-decoder baseline explains less than PCA, whereas on AlphaEarth it explains more and on Tessera about as much.

Because these embeddings are wider, the same $k$ explains much less of them.
At $k{=}16$, the SAE explains $0.65$ of the variance of OlmoEarth and $0.64$ of MMEarth, far less than on AlphaEarth ($0.97$), whereas at $k/D{=}0.25$ it explains $0.87$ and $0.81$.
A larger dictionary helps little: quadrupling $M$ at $k{=}16$ changes the FVE by at most $0.014$ on the validation split (one seed per setting).
For a new embedding, $k/D$ is therefore the first setting to choose.
This rule concerns reconstruction only, and matching $k/D$ leaves $M/D$ unmatched ($16$ for AlphaEarth, $5.3$ for OlmoEarth), so the table does not rank the embeddings.

\begin{table}[h] 
\centering
\caption{Fraction of variance explained on three further Earth embeddings, on the test split, for BatchTopK SAEs and the frozen-decoder baseline of the same $M$ and $k$, and for the first $L_0$ principal components and the $L_0$ raw dimensions of largest variance, with $L_0$ that of the SAE.
Mean over three seeds; $^\ast$one seed.}
\label{tab:supp_products_recon}
\small
\begin{tabular}{lrrrrrrrr}
\toprule
Embedding & $D$ & $M$ & $k$ & $L_0$ & SAE & frozen & PCA & raw \\
\midrule
Copernicus-FM & 768 & 1024 & 16  & 16  & 0.874 & 0.124 & 0.676 & 0.108 \\
              & 768 & 1024 & 192 & 185 & 0.952 & 0.490 & 0.931 & 0.431 \\
\midrule
OlmoEarth     & 768 & 4096 & 16  & 16  & 0.650$^\ast$ & 0.153$^\ast$ & 0.295 & 0.021 \\
              & 768 & 4096 & 192 & 185 & 0.870 & 0.598 & 0.785 & 0.241 \\
\midrule
MMEarth       & 320 & 4096 & 16  & 18  & 0.641$^\ast$ & 0.328$^\ast$ & 0.355 & 0.056 \\
              & 320 & 4096 & 80  & 84  & 0.809 & 0.639 & 0.697 & 0.263 \\
\bottomrule
\end{tabular}
\end{table}

\FloatBarrier
\clearpage
\section{Spatial scale}\label{sec:supp_spatial}

This appendix supports §\ref{sec:results_freq}.

\paragraph{Details of Figure~\ref{fig:variogram}.} The correlation length is defined in §\ref{sec:interpretability} and computed on the nested sample of Appendix~\ref{sec:supp_eval}; the frozen-decoder baseline (``frozen decoder'') is seed $0$ (every share of the other two seeds lies within $0.02$ of it).
For $13$ to $21\%$ of the SAE features, the correlation length falls in the unsampled band between $640$\,m and $13$\,km.

\subsection{Semivariance curves}\label{sec:supp_variogram_tessera}

\begin{figure}[ht]
    \centering
    \includegraphics[width=\linewidth]{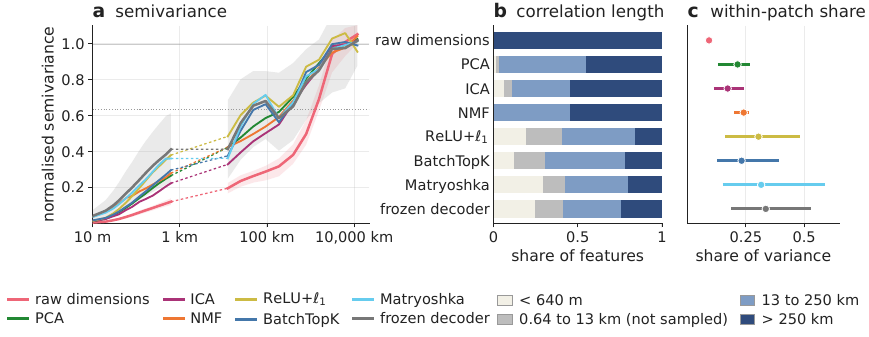}
    \caption{Spatial scales of the features on AlphaEarth. (a)~Median normalised semivariance against separation. (b)~Share of features by correlation length. (c)~Share of the variance of a feature within a patch (median and interquartile range over features).}
    \label{fig:supp_variogram_ae}
\end{figure}

Figure~\ref{fig:supp_variogram_ae} shows the spatial scales of the features on AlphaEarth in three panels.
Panel~a shows, for each decomposition, the median over its features of the normalised semivariance $\gamma_j(h)/\sigma_j^2$ against the separation $h$.
Its dotted line marks the level $1-1/e$ at which the correlation length is read off, and its dashed lines bound the band between $640$\,m and $13$\,km, which no pixel pair of the sample spans.
Its shaded bands are the interquartile ranges over the raw dimensions and over the features of the frozen-decoder baseline.
Panel~b repeats the share of features by correlation length from Figure~\ref{fig:variogram}, and panel~c shows the share of a feature's variance that lies within a $640$\,m patch (median and interquartile range over features).

\begin{figure}[ht]
    \centering
    \includegraphics[width=\linewidth]{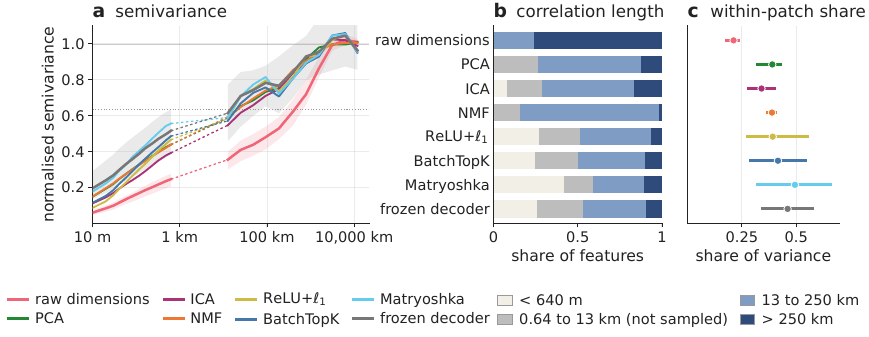}
    \caption{Spatial scales of the features on Tessera (cf.\ Figure~\ref{fig:supp_variogram_ae}).}
    \label{fig:supp_variogram_tessera}
\end{figure}

Figure~\ref{fig:supp_variogram_tessera} repeats the figure on Tessera.
In every decomposition, a larger share of the variance lies at fine scales than on AlphaEarth: $21\%$ of a raw dimension's variance lies within patches, against $34$ to $39\%$ for PCA, ICA and NMF and $39$, $42$ and $49\%$ for ReLU+$\ell_1$, BatchTopK and Matryoshka.
Again, only the raw dimensions vary mainly between distant places: $76\%$ of them have a correlation length beyond $250$\,km, and none below $640$\,m.
The frozen-decoder baseline is closest to Matryoshka, with $46\%$ of the variance within patches ($47\%$ for the other two seeds) and a nugget of $0.20$.
Among the SAEs, ReLU+$\ell_1$ has the smallest nugget ($0.09$, against $0.06$ for the raw dimensions).
For $27\%$ of the PCA, $21\%$ of the ICA and $16\%$ of the NMF components, the correlation length falls in the unsampled band between $640$\,m and $13$\,km.

\FloatBarrier
\subsection{Patch-mean Moran's $I$}\label{sec:supp_decay_tessera}

\begin{figure}[ht]
    \centering
    \includegraphics[width=\linewidth]{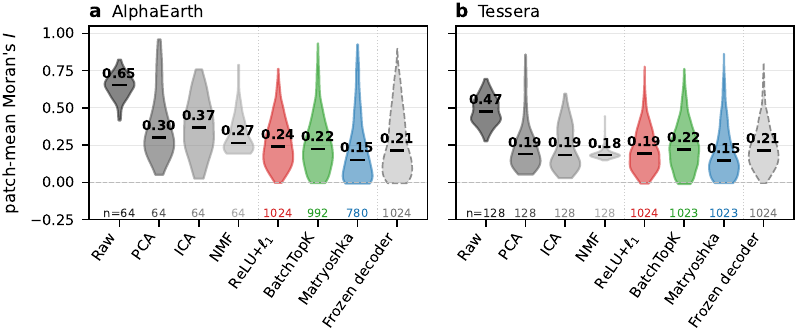}
    \caption{Per-feature patch-mean Moran's $I$ on (a)~AlphaEarth and (b)~Tessera; the median is printed above each violin and the number of features below it.
    The frozen-decoder baseline (seed $0$) lies within the range of the trained SAEs.}
    \label{fig:halflife_summary}
\end{figure}

Patch-mean Moran's $I$ (Eq.~\ref{eq:supp_moran}) measures, with one number per feature, how similar the mean activations of neighbouring patches are.
On AlphaEarth (Figure~\ref{fig:halflife_summary}a), the median Moran's $I$ of every SAE variant is lower than that of the raw embedding and the dense bases, with medians of $0.15$ to $0.24$ against $0.65$ (raw), $0.30$ (PCA), $0.37$ (ICA) and $0.27$ (NMF).
The frozen-decoder baseline reaches a median of $0.21$ ($0.21$ and $0.23$ for the other two seeds), against $0.22$ for the trained BatchTopK SAE.
On Tessera (Figure~\ref{fig:halflife_summary}b), the median of the raw embedding ($0.47$) is still higher than that of every SAE variant, but that of BatchTopK ($0.22$) now lies above PCA ($0.19$), ICA ($0.19$) and NMF ($0.18$), and the frozen-decoder baseline reaches $0.21$ ($0.22$ and $0.20$ for the other two seeds).
On both embeddings, Moran's $I$ therefore does not distinguish SAEs with a trained decoder from the frozen-decoder baseline.

\FloatBarrier
\subsection{Dominant features}\label{sec:supp_dominant}

\begin{figure}[t]
    \centering
    \includegraphics[width=\linewidth]{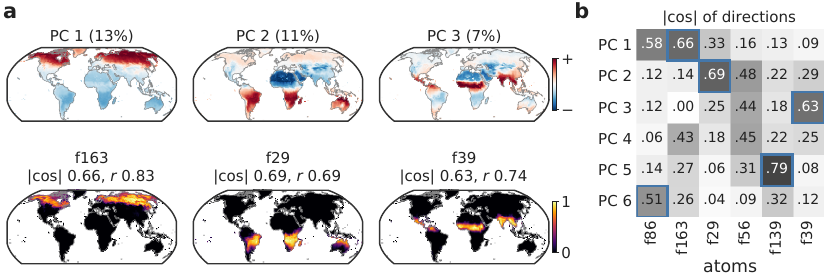}
    \caption{High-energy SAE features point along the leading principal components. (a)~Scores of the first three principal components of AlphaEarth (top, each oriented to correlate positively with the feature below) and activation of the Matryoshka feature whose atom is closest (bottom), with the absolute cosine of component and atom and the correlation $r$ of score and activation. (b)~Absolute cosine between the first six components and the atoms of the six features with the largest energy share; the best match among all atoms is outlined.}
    \label{fig:pcs_atoms}
\end{figure}

One reason why SAE features vary on shorter scales than principal components (§\ref{sec:results_freq}) is that the SAE splits the leading components into pieces.
Figure~\ref{fig:pcs_atoms} compares the leading principal components of AlphaEarth with the atoms of the Matryoshka features of largest energy share (Appendix~\ref{sec:supp_energy}).
The six features of largest energy share together carry $40\%$ of the energy, and the wet-tropical feature f139 of Figure~\ref{fig:qualitative} is one of them.
For five of the six leading principal components, the closest of all $1024$ atoms belongs to one of these six features (panel b), at an absolute cosine of $0.51$ to $0.79$.
By comparison, the closest of $1024$ random unit vectors in $64$ dimensions reaches about $0.41$ ($0.48$ at the $95$th percentile).
For the first three components, the activation of that feature also correlates with the component score, at $r=0.69$ to $0.83$ over pixels (panel a).
Because an activation is non-negative, one feature covers only one end of a component, as the boreal feature f163 does for the first component.
The leading components vary mainly between regions, so such a feature mainly encodes where a pixel is, like the features of Figure~\ref{fig:overview}c.
A plausible reason why the largest features follow the leading components is the training objective: an SAE is trained to reconstruct, so it favours directions of large variance, and on AlphaEarth $91\%$ of the variance lies between $640$\,m patches.
For interpretation, this means that the features that dominate the code mainly encode location, and that a concept that occurs on several continents may be split over several regional features, such as the three woody-vegetation features of Figure~\ref{fig:qualitative}.

\paragraph{Details of Figure~\ref{fig:pcs_atoms}.} The maps show means over the training patches in each $2^\circ$ cell, and the atoms are those of the Matryoshka SAE closest to each component.
Above each map, we report the absolute cosine between component and atom and the correlation $r$ of score and activation over pixels; each activation is scaled to its maximum.

\FloatBarrier
\clearpage
\section{Concept alignment}\label{sec:supp_alignment}

This appendix supports §\ref{sec:results_emf}.

\paragraph{Details of Figure~\ref{fig:monosemanticity}.} Panel a counts, for every live feature, the concept groups with which it is aligned (Appendix~\ref{sec:supp_eval}), and also shows a kernel density of the counts, drawn without its own axis.
The grey series is the frozen-decoder baseline ($M{=}1024$, $k{=}16$), seed $0$, with $974$ live features; the other two seeds give nearly the same shares ($88$ and $89\%$ at zero or one group).
Panel a counts all $20$ groups.
Panels b and c show the $13$ groups whose best Matryoshka feature reaches the highest folded AUROC, in the order of Table~\ref{tab:supp_concept_groups}.
Summit and ridge, the two geomorphon groups on which no feature or raw dimension reaches a folded AUROC of $0.7$, are not candidates for these panels.
Column $i$ of each panel is the best aligned and most monosemantic feature (b) or raw dimension (c) for the group of row $i$.
Among the features whose folded AUROC on that group reaches $0.65$, it is the one aligned with the fewest groups, with ties broken by the higher AUROC on the group.
A group without an aligned feature takes the feature of highest AUROC.
No feature appears twice, because the assignment is solved jointly for all rows by the Hungarian algorithm.
Panels b and c share one grey scale that steps from light to dark at the alignment threshold, so that aligned cells (folded AUROC $\geq 0.65$) are dark; colour carries no other meaning.
Figures~\ref{fig:supp_mono_tessera} and~\ref{fig:supp_batchtopk_mono} use the same rule on their own SAE.
Concept groups co-occur on the ground (Appendix~\ref{sec:supp_concept_groups}), so some alignment off the diagonal of panels b and c is real.
A feature aligned with no group has not shown that it is monosemantic: it may represent a concept that the registry lacks, or none that a reader would name.

\subsection{Monosemanticity on Tessera}\label{sec:supp_mono_tessera}

\begin{figure}[ht]
    \centering
    \includegraphics[width=\linewidth]{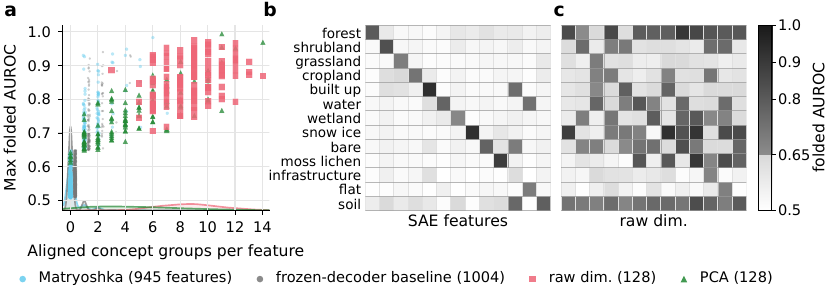}
    \caption{Monosemanticity on Tessera (cf.\ Figure~\ref{fig:monosemanticity}).
    The panels compare a Matryoshka SAE on Tessera ($M{=}1024$, $k{=}16$) with the raw Tessera dimensions ($D{=}128$) and the PCA-$128$ components.
    Raw Tessera dimensions align with a median of $9$ concept groups and PCA-$128$ components with $4$.
    Of the $945$ live Matryoshka features, $93\%$ align with at most one group, and so do $97\%$ of the $3{,}010$ live features of three seeds of the frozen-decoder baseline (grey).}
    \label{fig:supp_mono_tessera}
\end{figure}

Figure~\ref{fig:supp_mono_tessera} repeats the three panels of Figure~\ref{fig:monosemanticity} on Tessera: it compares a Matryoshka SAE trained on Tessera with the raw Tessera dimensions and the PCA-$128$ components.
A raw Tessera dimension aligns with a median of $9$ concept groups and a PCA-$128$ component with $4$, whereas a Matryoshka feature aligns with a median of zero.
As on AlphaEarth, the heatmap of SAE features is diagonal, whereas that of raw dimensions has many aligned cells off the diagonal.
The frozen-decoder baseline scores as well on this count: $97\%$ of its $1{,}004$ live features align with at most one concept group, against $93\%$ of the $945$ live Matryoshka features.
On Tessera, too, the contrast with the dense bases thus follows from sparsity rather than from the learned dictionary.

\FloatBarrier
\subsection{Monosemanticity for BatchTopK}\label{sec:supp_batchtopk_mono}

\begin{figure}[ht]
    \centering
    \includegraphics[width=\linewidth]{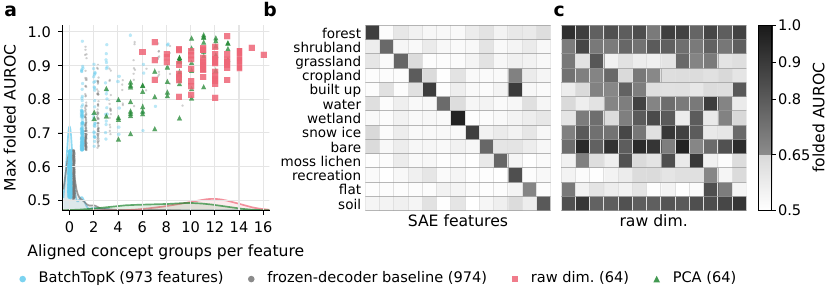}
    \caption{Monosemanticity of BatchTopK SAE features (cf.\ Figure~\ref{fig:monosemanticity}).
    Same panels, with BatchTopK in place of Matryoshka ($M{=}1024$, $k{=}16$, AlphaEarth).
    Of the $973$ live BatchTopK features, $94\%$ align with at most one concept group, against $89\%$ of the live features of the frozen-decoder baseline (grey, seed $0$).}
    \label{fig:supp_batchtopk_mono}
\end{figure}

Figure~\ref{fig:supp_batchtopk_mono} repeats the three panels of Figure~\ref{fig:monosemanticity} with a BatchTopK SAE in place of the Matryoshka SAE, at the same width and sparsity ($M{=}1024$, $k{=}16$).
Of the $973$ live BatchTopK features, $94\%$ align with at most one concept group, against $89\%$ of the live features of the frozen-decoder baseline (seed $0$; $88\%$ and $89\%$ for the other two seeds).
The heatmap of SAE features is as diagonal as for Matryoshka.

\FloatBarrier
\subsection{Further Earth embeddings}\label{sec:supp_mono_products}

We repeat the alignment count of §\ref{sec:results_emf} on the $80$\,m tokens of OlmoEarth and MMEarth at $k/D{=}0.25$ (Appendix~\ref{sec:supp_setup_products}), with each token labelled by the majority class of its pixels (Table~\ref{tab:supp_products_mono}).
The pattern of AlphaEarth holds.
A raw dimension is aligned with a median of $7$ concept groups on OlmoEarth and $6$ on MMEarth, and almost none with at most one, whereas $94\%$ and $98\%$ of the live SAE features are aligned with at most one group, most of them with none.
The frozen-decoder baseline is as specific ($92\%$ and $98\%$) and has more aligned features than the SAE ($995$ against $609$ on OlmoEarth, $473$ against $265$ on MMEarth), as on AlphaEarth.
At $k{=}16$ (one seed), almost every live feature is aligned with no group, for the SAE and the baseline alike ($98$ to $100\%$).

\begin{table}[h]
\centering
\caption{Concept alignment on the $80$\,m tokens of OlmoEarth and MMEarth ($4{,}000$ test patches): number of live features, share aligned with at most one and with exactly one concept group, number aligned with at least one, and the median number of groups per feature.
SAE and frozen-decoder baseline at $k/D{=}0.25$ ($M{=}4096$), mean over three seeds.}
\label{tab:supp_products_mono}
\small
\begin{tabular}{llrrrrr}
\toprule
Embedding & Representation & live & $\leq 1$ & $=1$ & $\geq 1$ & median \\
\midrule
OlmoEarth & raw dimensions & 768  & 0.01 & 0.01 & 767 & 7 \\
          & PCA            & 768  & 0.35 & 0.25 & 698 & 2 \\
          & SAE            & 4095 & 0.94 & 0.09 & 609 & 0 \\
          & frozen decoder & 4096 & 0.92 & 0.16 & 995 & 0 \\
\midrule
MMEarth   & raw dimensions & 320  & 0.00 & 0.00 & 320 & 6 \\
          & PCA            & 320  & 0.17 & 0.12 & 306 & 3 \\
          & SAE            & 3914 & 0.98 & 0.05 & 265 & 0 \\
          & frozen decoder & 4002 & 0.98 & 0.10 & 473 & 0 \\
\bottomrule
\end{tabular}
\end{table}

\FloatBarrier
\subsection{Concept coverage}\label{sec:supp_alignment_scores}

The alignment counts of §\ref{sec:results_emf} ask how specific each feature is.
A complementary question is how many reference classes a representation covers at all.
The unique coverage UCov~\cite{karvonen.etal_2025} counts the product classes on which at least one feature reaches a folded AUROC (Eq.~\ref{eq:folded_auroc}) of $\tau$,
\begin{equation}
    \mathrm{UCov}_\tau \;=\; \big|\{c : \max_j \mathrm{AUROC}^{\pm}(z_j; c) \geq \tau\}\big|, \label{eq:ucov_emf}
\end{equation}
counted over the $159$ categorical product classes that occur often enough in the AlphaEarth evaluation sample to be scored.
Coverage must be computed with the same statistic for every representation.
For instance, crediting each feature only with its single best-matching class would cap the count of a dense basis at its $D$ features, and would give the SAEs an advantage that disappears under a common statistic.

Table~\ref{tab:ucov_one_statistic} reports UCov for every representation.
On AlphaEarth, the raw embedding ($151$ of $159$ classes at $\tau{=}0.60$) and PCA ($153$) cover more classes than every SAE variant ($111$ to $146$), so coverage does not rank any SAE above the raw embedding.
It does rank the trained BatchTopK SAE above the frozen-decoder baseline of the same dictionary size and sparsity ($140$ against $129$ at $\tau{=}0.60$, $116$ against $110$ at $\tau{=}0.65$), and the frozen-decoder baseline above an untrained random dictionary ($120$ and $88$), whose encoder is the transpose of its random decoder.
Each of these is a single training run, so the margins come without an estimate of their spread over seeds.

We do not use the effective monosemanticity EMF~\cite{karvonen.etal_2025}, the sum over aligned features of one over the number of concept groups each is aligned with.
It does not separate a trained from a frozen decoder: at $M{=}1024$, the frozen-decoder baseline scores $167.4$ and a trained BatchTopK SAE $149.1$.
It also grows with dictionary width even for an untrained random dictionary, from $16.3$ at $M{=}64$ to $133.4$ at $M{=}1024$.

\begin{table}[h]
\centering
\footnotesize
\caption{UCov, computed with the same statistic for every representation: the number of product classes that at least one feature aligns with at $\mathrm{AUROC}^{\pm}\geq\tau$, with the SAEs scored on their sparse codes, at $M{=}1024$, $k{=}16$.
On AlphaEarth, $159$ classes are scored, and every row uses the same pixel sample.
The frozen decoder and the untrained random dictionary have the dictionary size and sparsity of the SAEs.
Every row is one run; on Tessera, these two controls were not scored.}
\label{tab:ucov_one_statistic}
\begin{tabular}{lcccc}
\toprule
 & \multicolumn{2}{c}{AlphaEarth} & \multicolumn{2}{c}{Tessera} \\
\cmidrule(lr){2-3}\cmidrule(lr){4-5}
 & $\tau{=}0.60$ & $\tau{=}0.65$ & $\tau{=}0.60$ & $\tau{=}0.65$ \\
\midrule
Raw                       & $151$ & $147$ & $133$ & $127$ \\
PCA-$D$                   & $153$ & $143$ & $132$ & $123$ \\
ICA-$D$                   & $149$ & $128$ & $115$ & $94$ \\
NMF-$D$                   & $139$ & $114$ & $131$ & $122$ \\
\midrule
ReLU+$\ell_1$             & $111$ & $87$  & $74$  & $45$ \\
BatchTopK                 & $140$ & $116$ & $111$ & $80$ \\
Matryoshka                & $146$ & $126$ & $117$ & $98$ \\
\midrule
Frozen decoder            & $129$ & $110$ & --  & -- \\
Untrained random dictionary & $120$ & $88$ & -- & -- \\
\bottomrule
\end{tabular}
\end{table}

\FloatBarrier
\subsection{Monosemanticity scores of Betti et al.}\label{app:betti_scores}

Betti et al.~\cite{betti.etal_2026} score the features of SAEs on location encoders with the visual monosemanticity score (MS) of Pach et al.~\cite{pach.etal_2025} and with a geographic variant (GeoMS).
We evaluate both on our SAEs.
Both scores ask whether the samples on which a feature is strongly active resemble each other, in their images for MS and in their location for GeoMS.
Following the released code of both papers, the activations $a_{ij}$ of feature $j$ on the samples $i=1,\dots,N$ are min-max scaled to $\tilde a_{ij}\in[0,1]$, and
\begin{equation}
  \mathrm{MS}_j \;=\; \frac{\sum_{i<i'}\tilde a_{ij}\,\tilde a_{i'j}\,s_{ii'}}{\sum_{i<i'}\tilde a_{ij}\,\tilde a_{i'j}},
\end{equation}
where the similarity $s_{ii'}$ of samples $i$ and $i'$ is the cosine similarity of their image embeddings for MS, and $s_{ii'}=\exp(-d_{ii'}/500\,\mathrm{km})$ with their great-circle (Haversine) distance $d_{ii'}$ for GeoMS.
The sums run over all pairs of samples, and features active on fewer than $20$ samples are not scored.
The score is a weighted mean of similarities, not a correlation, so a feature that is active uniformly scores the mean pairwise similarity of the sample, which we call the uniform level.

\paragraph{Implementation choices.} Betti et al.\ score point locations.
Ours are $640$\,m patches, each represented by the mean activation over its pixels and located at its centroid.
We keep the length scale of $500$\,km from their code.
GeoMS uses the $1{,}500$ test patches of Figure~\ref{fig:halflife_summary}.
Sentinel-2 images exist for only $314$ of these, so MS uses $3{,}000$ other test patches that have one, drawn at random.
The image is the Sentinel-2 L2A crop of the patch footprint, embedded with the image tower of SatCLIP~\cite{klemmer.etal_2025a}, which Betti et al.\ use for Sentinel-2.
These choices were fixed before the scores were computed.
The SAE rows are the models of Table~\ref{tab:eight_criteria}, and on AlphaEarth we add three further BatchTopK seeds.

\paragraph{Result.} Table~\ref{tab:betti_scores} reports the mean and the maximum of each score over features.
Raw dimensions and PCA and ICA components score at the uniform level on both MS and GeoMS.
On the mean over features, every trained SAE scores above all three seeds of the frozen-decoder baseline, for both scores and on both embeddings.
On AlphaEarth, each of the three further BatchTopK seeds also scores above each frozen seed.
The frozen decoder already reaches much of the trained score.
On AlphaEarth it reaches a mean GeoMS of $0.102$ against a uniform level of $0.008$, and a mean MS of $0.437$ against $0.262$.
Measured above the uniform level, this is $69$--$81\%$ (GeoMS) and $83$--$87\%$ (MS) of the score of the trained SAEs.
On Tessera these shares are $50$--$63\%$ (GeoMS) and $64$--$66\%$ (MS).
On the mean over features, both scores thus separate trained from random dictionaries, by a small margin on AlphaEarth and a larger one on Tessera.
A large part of their value, however, comes from the sparse code itself.
The maximum over features does not separate trained from frozen reliably.
For MS, ReLU+$\ell_1$ lies within the range of the frozen seeds on both embeddings, and the three trained BatchTopK seeds overlap this range on AlphaEarth.
For GeoMS, the AlphaEarth Matryoshka SAE lies within this range.
The means are also taken over different numbers of features, because features active on fewer than $20$ patches are dropped.
On the GeoMS patches of AlphaEarth, $464$ to $1{,}022$ of the $1{,}024$ features are scored for the trained SAEs and $878$ to $893$ for the frozen decoder.
We therefore suggest reporting these scores together with the uniform level and a frozen-decoder baseline.

\begin{table}[h]
  \centering
  \caption{Monosemanticity scores of Betti et al.~\cite{betti.etal_2026} on our dictionaries: mean and maximum over scored features.
  GeoMS on $1{,}500$ test patches, MS on $3{,}000$ test patches with a Sentinel-2 image.
  The row \emph{Uniform} is the score of a feature that fires equally everywhere.
  Frozen decoder and trained BatchTopK seeds: mean $\pm$ standard deviation over three seeds; ICA: mean over three fits.
  The maximum is comparable only between dictionaries of equal width (all SAEs have $1{,}024$ features).}
  \label{tab:betti_scores}
  \scriptsize
  \setlength{\tabcolsep}{2.5pt}
  \resizebox{\linewidth}{!}{%
  \begin{tabular}{lcccccccc}
    \toprule
    & \multicolumn{4}{c}{AlphaEarth} & \multicolumn{4}{c}{Tessera} \\
    \cmidrule(lr){2-5}\cmidrule(lr){6-9}
    & \multicolumn{2}{c}{GeoMS} & \multicolumn{2}{c}{MS} & \multicolumn{2}{c}{GeoMS} & \multicolumn{2}{c}{MS} \\
    & mean & max & mean & max & mean & max & mean & max \\
    \midrule
    Uniform                  & $0.008$ & --      & $0.262$ & --      & $0.008$ & --      & $0.262$ & --      \\
    Raw                      & $0.009$ & $0.010$ & $0.265$ & $0.281$ & $0.009$ & $0.010$ & $0.263$ & $0.280$ \\
    PCA                      & $0.009$ & $0.010$ & $0.262$ & $0.282$ & $0.008$ & $0.010$ & $0.262$ & $0.282$ \\
    ICA                      & $0.009$ & $0.010$ & $0.262$ & $0.270$ & $0.008$ & $0.009$ & $0.262$ & $0.269$ \\
    \midrule
    Frozen decoder           & $0.102{\scriptstyle\pm0.001}$ & $0.453{\scriptstyle\pm0.046}$ & $0.437{\scriptstyle\pm0.002}$ & $0.800{\scriptstyle\pm0.009}$ & $0.063{\scriptstyle\pm0.001}$ & $0.390{\scriptstyle\pm0.044}$ & $0.392{\scriptstyle\pm0.002}$ & $0.767{\scriptstyle\pm0.034}$ \\
    ReLU+$\ell_1$            & $0.125$ & $0.547$ & $0.472$ & $0.807$ & $0.095$ & $0.580$ & $0.464$ & $0.798$ \\
    BatchTopK                & $0.140$ & $0.611$ & $0.465$ & $0.807$ & $0.097$ & $0.477$ & $0.460$ & $0.824$ \\
    Matryoshka               & $0.134$ & $0.488$ & $0.463$ & $0.847$ & $0.118$ & $0.623$ & $0.466$ & $0.840$ \\
    BatchTopK, 3 seeds       & $0.144{\scriptstyle\pm0.004}$ & $0.570{\scriptstyle\pm0.052}$ & $0.467{\scriptstyle\pm0.005}$ & $0.824{\scriptstyle\pm0.019}$ & -- & -- & -- & -- \\
    \bottomrule
  \end{tabular}}
\end{table}

\FloatBarrier
\clearpage
\section{Aligned features and sub-concepts}\label{sec:supp_subconcepts}

This appendix supports §\ref{sec:results_qualitative}.

\paragraph{Details of Figure~\ref{fig:qualitative}.} Panel a shows $640$\,m patches at $10$\,m resolution on AlphaEarth; $-$ marks a sign-flipped raw dimension, and each feature row shares one colour scale.
The patches are at, in order, for water $57^\circ$N $158^\circ$W, $49^\circ$N $119^\circ$E, $60^\circ$N $121^\circ$W (with the class) and $1^\circ$S $67^\circ$W, $20^\circ$S $136^\circ$E, $55^\circ$N $37^\circ$E (without), and for built-up $26^\circ$N $85^\circ$E, $32^\circ$N $120^\circ$E, $50^\circ$N $7^\circ$E and $40^\circ$N $103^\circ$W, $3^\circ$S $75^\circ$W, $35^\circ$N $91^\circ$W.
Features of the frozen-decoder baseline can align with these classes as well (Figure~\ref{fig:monosemanticity}a), so panel a illustrates what a sparse feature looks like and is not evidence of what the dictionary has learned.
Panel c shows the $2{,}000$ test patches on which each feature is most active.
Panels d to f show the $2^\circ$ cells whose majority class in each map belongs to the classes that dominate the feature's patches (Table~\ref{tab:supp_subconcepts}).

Sub-concept discovery asks whether several SAE features divide one coarse land-cover class into finer types.
We treat it as one possible use of the sparse code and show it qualitatively.
The most useful divisions are those that no global map contains yet, but for them no independent reference exists to score them against.
For the woody-vegetation example of Figure~\ref{fig:qualitative}b--f, three independent maps do exist, and we compare against them descriptively.

\paragraph{Setup.} We take the three features of the Matryoshka SAE on AlphaEarth shown in Figure~\ref{fig:qualitative}b--f (f139, f124, f35) and, for each, the $2{,}000$ of the $99{,}935$ test patches with the highest mean activation over the patch's $64{\times}64$ pixels.
Each patch is labelled with its majority class in three products that are in neither the reference registry nor the concept taxonomy: Köppen-Geiger climate zones from CHELSA V2.1~\cite{karger.etal_2017, peel.etal_2007}, RESOLVE biomes~\cite{dinerstein.etal_2017}, and potential natural vegetation (PNV)~\cite{hengl.etal_2018}.
PNV predicts the vegetation that would grow without human land use, so it is a counterfactual rather than an observation.
For each class we report its share of a feature's $2{,}000$ patches and its lift, the share divided by the class's share among all test patches.
Against ESA WorldCover at patch level, f139 and f35 align with tree cover (folded AUROC $0.689$ and $0.683$), and f124 comes closest to shrubland ($0.641$, below the $0.65$ alignment threshold).

\paragraph{Result.} Each feature concentrates on a distinct climate zone and biome (Table~\ref{tab:supp_subconcepts}, Figure~\ref{fig:qualitative}b--f).
f139 covers the wet tropics: $80\%$ of its patches lie in tropical rainforest climate (Af, lift $15.6$), $98\%$ in the tropical moist broadleaf forest biome, and nearly all of them ($1{,}999$) in closed evergreen broadleaf PNV.
f124 covers dry savanna and shrubland: hot semi-arid (BSh, $39\%$) and tropical savanna (Aw, $24\%$) climates, the tropical grassland, savanna and shrubland biome ($59\%$), and PNV shrubs ($69\%$).
f35 covers boreal and conifer forest: subarctic and warm-summer continental climates (Dfc and Dfb, $91\%$ together), the boreal ($55\%$) and temperate conifer ($23\%$, lift $8.0$) forest biomes, and closed evergreen needleleaf PNV ($65\%$, lift $8.6$).
The mean absolute latitudes of the three sets of patches are $3.0^\circ$, $23.7^\circ$ and $55.3^\circ$.

\paragraph{What the comparison does not show.} The agreement is descriptive: it is measured on each feature's own most active patches, without correcting for the choice of the best-matching class and without a null model for geography.
A stricter test, run on a partition of all forest patches rather than on each feature's most active patches, does not support more than this.
It assigns every forest patch (a patch with ESA WorldCover tree cover on at least half of its pixels) to one of up to six aligned features, the one whose patch-mean activation, divided by its mean active level, is largest.
It then compares the lift of each resulting sub-concept with that of size-matched, spatially compact random partitions of the same patches, of which we draw $50$, correcting for the choice among sub-concepts.
The forest partition cannot be distinguished from these random partitions (family-wise $p=0.14$ against PNV and $p=0.31$ against RESOLVE biomes).
Climate and biome maps are spatially smooth, so any compact subdivision of a class agrees with them to a similar degree.
The three features are climatically and geographically coherent, but this evidence does not separate them from a geographic split of forest, and we do not claim that the SAE recovers a biome taxonomy.

\begin{table}[ht]
    \centering
    \caption{Agreement of the three features of Figure~\ref{fig:qualitative}b--f with independent climate and biome products, on each feature's $2{,}000$ most active test patches. Each cell lists the dominant classes, each with its share of the patches and, in brackets, its lift over the class's share among all test patches. The column $\overline{|\text{lat}|}$ is the mean absolute latitude of the patches.}
    \label{tab:supp_subconcepts}
    \small
    \begin{tabularx}{\linewidth}{@{}l r X X X@{}}
        \toprule
        Feature & $\overline{|\text{lat}|}$ & Köppen-Geiger & RESOLVE biome & PNV \\
        \midrule
        f139 & $3.0^\circ$ & Af 0.80 (15.6); Am 0.16 (4.2) & Trop.\ moist broadleaf 0.98 (6.7) & Closed evergr.\ broadleaf 1.00 (5.5) \\
        f124 & $23.7^\circ$ & BSh 0.39 (6.3); Aw 0.24 (1.9) & Trop.\ grassland, savanna, shrub 0.59 (3.7); temp.\ grassland 0.17 (2.1) & Shrubs 0.69 (2.8) \\
        f35 & $55.3^\circ$ & Dfc 0.61 (4.6); Dfb 0.30 (4.4) & Boreal forest 0.55 (4.8); temp.\ conifer 0.23 (8.0) & Closed evergr.\ needleleaf 0.65 (8.6) \\
        \bottomrule
    \end{tabularx}
\end{table}

\subsection{Sub-concepts of the frozen-decoder baseline}\label{sec:supp_subconcepts_frozen}

\begin{figure}[ht]
    \centering
    \includegraphics[width=\linewidth]{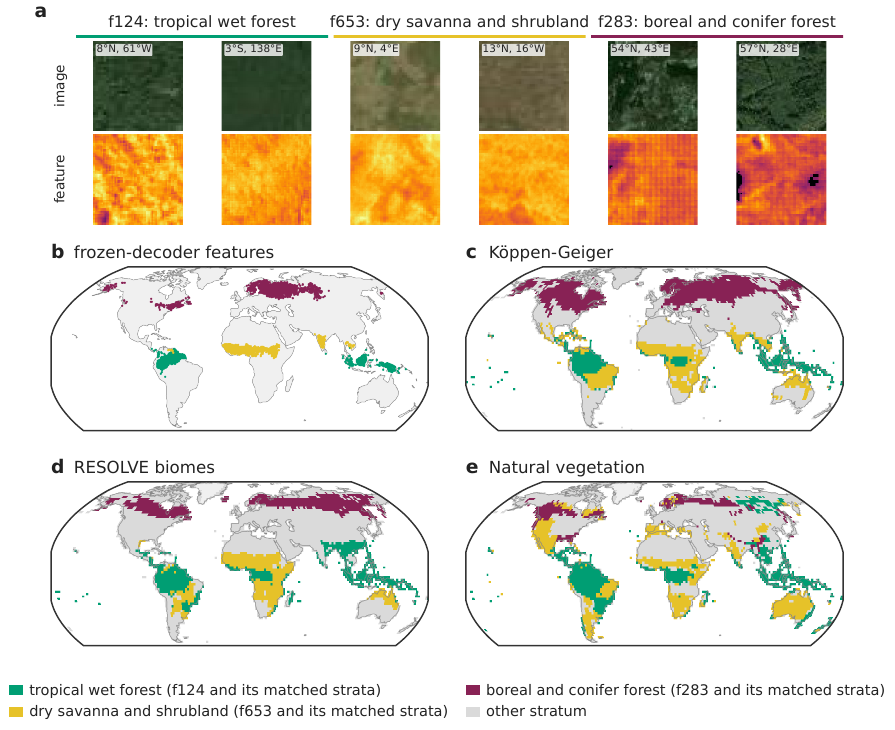}
    \caption{Woody-vegetation features of the frozen-decoder baseline on AlphaEarth, in the layout of Figure~\ref{fig:qualitative}b--f. (a)~Two top patches per feature, with imagery and activation. (b)~The $2{,}000$ most active test patches of each feature. (c--e)~The matched strata of Köppen-Geiger climate zones (Af+Am, BSh+Aw, Dfc+Dfb), RESOLVE biomes (tropical moist broadleaf, tropical grassland and savanna, boreal forest) and potential natural vegetation (evergreen broadleaf, shrubs, evergreen needleleaf).}
    \label{fig:supp_subconcepts_frozen}
\end{figure}

To test whether such features need a learned dictionary, we apply the same analysis to the frozen-decoder baseline (seed $0$).
A feature counts as woody if its folded AUROC reaches $0.65$ on a forest, shrubland or savanna class and its unfolded AUROC $A_{jc}$ on that class exceeds $0.5$, so that it is more active on the class than elsewhere.
For each Köppen-Geiger pair of Figure~\ref{fig:qualitative} (Af+Am, BSh+Aw and Dfc+Dfb), we take, in each SAE, the woody feature whose $2{,}000$ most active test patches lie most in that pair.
The baseline has $84$ woody features, against $47$ for the Matryoshka SAE.
Its best features place $99$, $98$ and $99\%$ of their patches in the three pairs, against $96$, $93$ and $96\%$ for the best Matryoshka features, and $10$, $11$ and $10$ of its woody features place at least $80\%$ there, against $2$, $3$ and $2$.
Concentration on one climate pair alone is therefore no evidence that a dictionary has learned a sub-concept.
The maps show a difference that this number misses (Figure~\ref{fig:supp_subconcepts_frozen}).
The frozen-decoder features cover only part of each stratum: the dry-savanna feature stays in the Sahel and misses southern Africa, South America and Australia, and the tropical feature misses the Congo basin.
The trained features of Figure~\ref{fig:qualitative}c cover much more of their strata.
The activation of the frozen-decoder features is also high on almost every pixel of their top patches, so they mark where a patch lies rather than structure inside it, whereas the trained features vary within a patch.
The trained features thus follow the climate and biome strata more completely, which a measure of coverage rather than concentration would have to quantify.

\FloatBarrier
\subsection{Galleries of aligned features}\label{sec:supp_qual_only}\label{sec:supp_qual_tessera}

\begin{figure}[p]
    \centering
    \includegraphics[width=\linewidth,height=0.95\textheight,keepaspectratio]{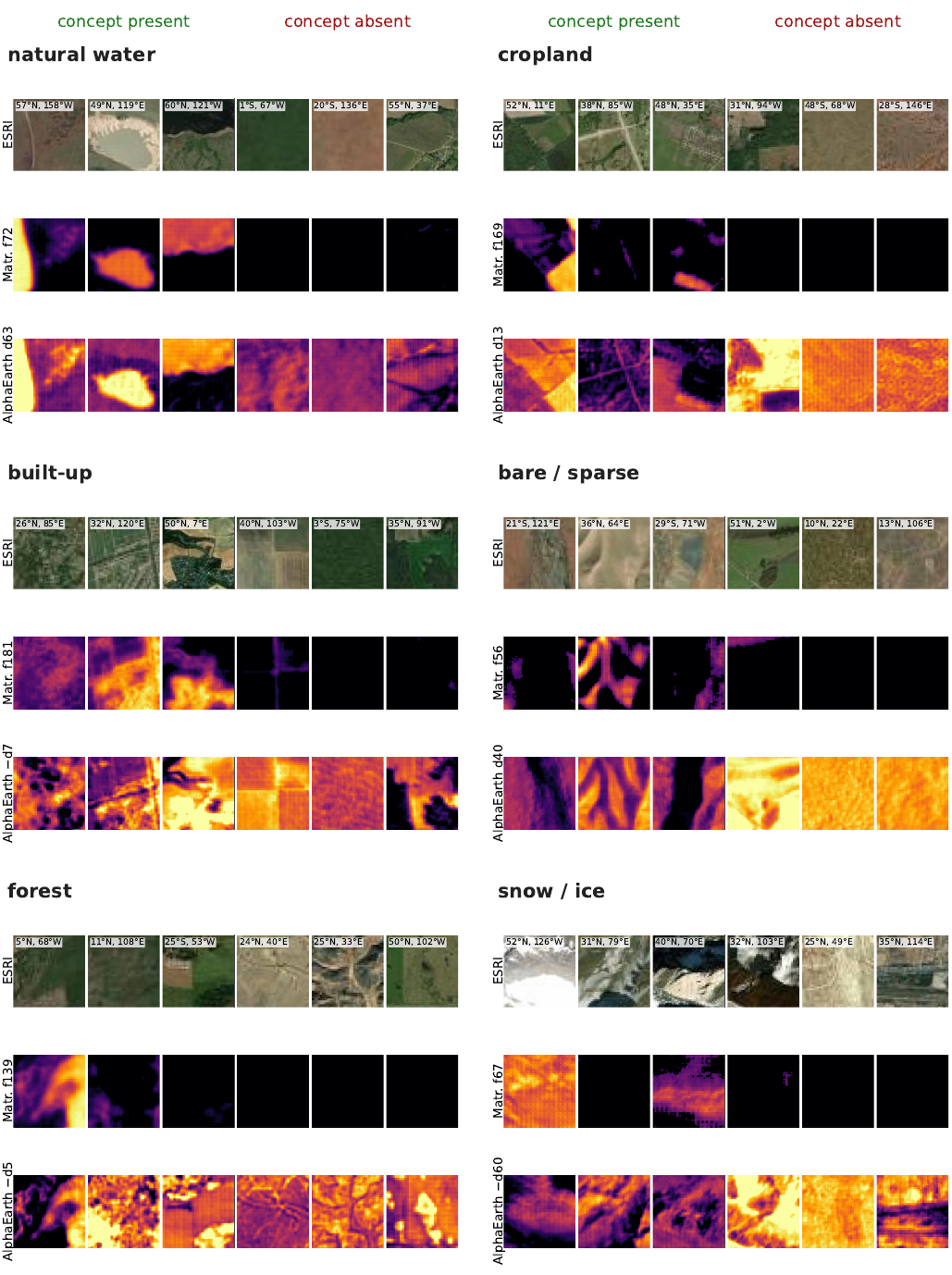}
    \caption{Qualitative gallery in the style of Figure~\ref{fig:qualitative}a for six registry concepts (natural water, built-up, forest, cropland, bare / sparse, snow / ice).
    Each block has three rows: Esri World Imagery, the activation of the Matryoshka SAE feature with the highest AUROC for the concept, and the best-aligned raw AlphaEarth dimension.
    The left three columns are concept-present patches with partial cover (e.g.\ river banks, lake edges, mixed urban land); the right three are concept-absent.
    For most concepts the SAE row is active only on the concept-present patches, whereas the raw row is active on both, illustrating the polysemanticity of the dense embedding.}
    \label{fig:supp_qual_only}
\end{figure}

\begin{figure}[p]
    \centering
    \includegraphics[width=\linewidth,height=0.95\textheight,keepaspectratio]{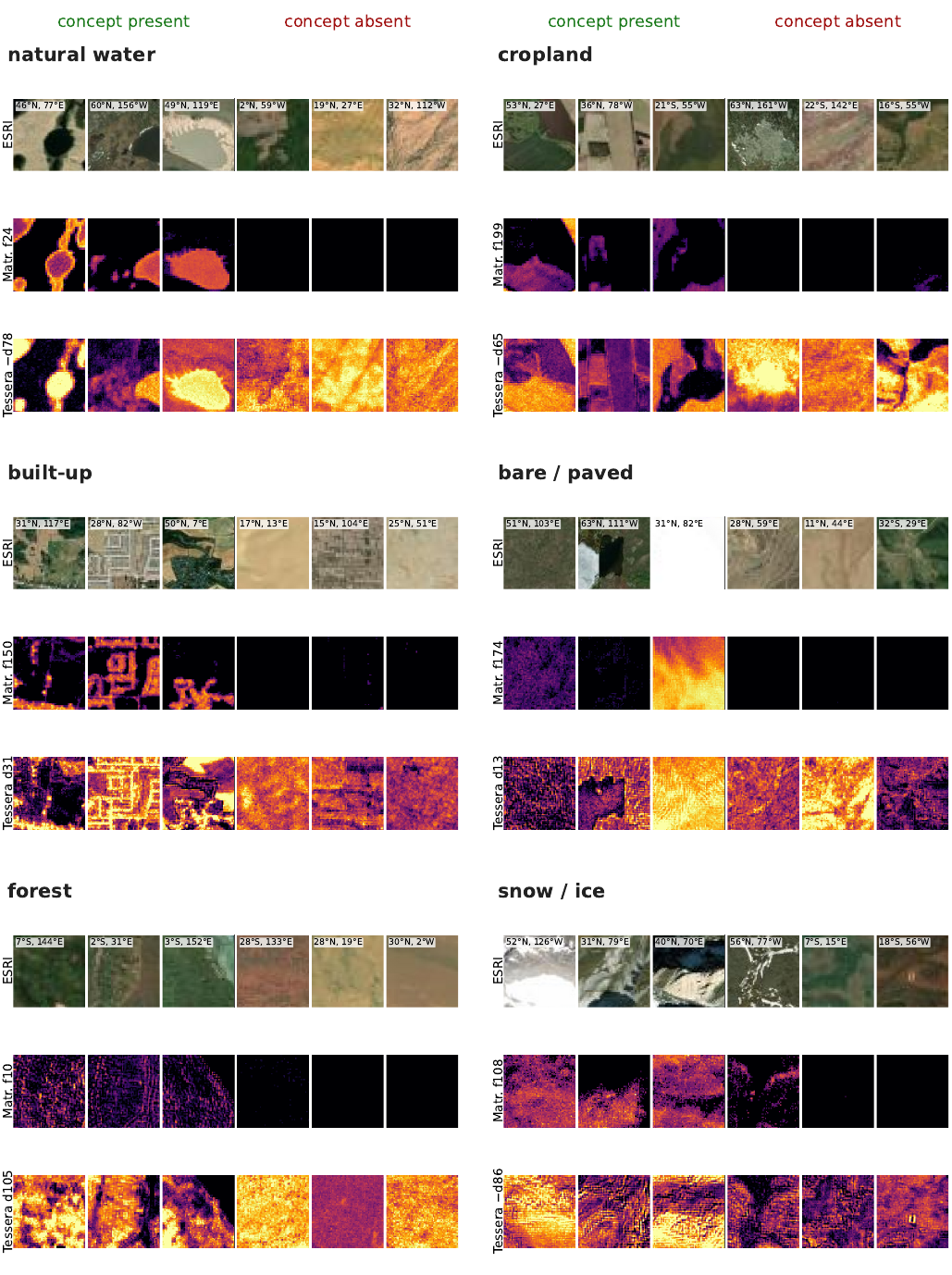}
    \caption{Qualitative gallery on Tessera (cf.\ Figure~\ref{fig:supp_qual_only}).
    Six registry concepts; each block has three rows (Esri imagery, Matryoshka SAE feature on Tessera, best-aligned raw Tessera dimension).
    The left three columns show patches that partly contain the concept, and the right three show patches without it.}
    \label{fig:supp_qual_tessera}
\end{figure}

Figures~\ref{fig:supp_qual_only} and~\ref{fig:supp_qual_tessera} extend the gallery of Figure~\ref{fig:qualitative}a to six concepts on AlphaEarth and on Tessera.
For most concepts, the best-aligned Matryoshka feature is active on the patches that contain the concept and silent on the others, whereas the best-aligned raw dimension is active on both.
For some concepts, such as forest and cropland on AlphaEarth, the feature covers only part of the class.

\FloatBarrier
\clearpage
\section{Area of applicability}\label{sec:supp_aoa_results}

This appendix extends the area-of-applicability (AoA) results of §\ref{sec:results_aoa} and Figure~\ref{fig:aoa_main}, whose setup §\ref{sec:downstream} describes.
Its outcome is mixed.
The distance in the SAE code ranks the errors of a canopy-height model better than the distance in the raw embedding in most hold-outs, but it does not do so for an elevation model.
On AlphaEarth, an untrained random sparse code of the same dictionary size and sparsity gains as much as a trained SAE, so the gain cannot be attributed to learned features.
We describe what holds, what does not, and why we think the two targets differ.
Throughout, the \emph{head} is the ridge regressor that predicts the target from a representation, either the raw embedding or an SAE code (Appendix~\ref{sec:supp_aoa}).

\begin{figure}[ht]
    \centering
    \includegraphics[width=\linewidth]{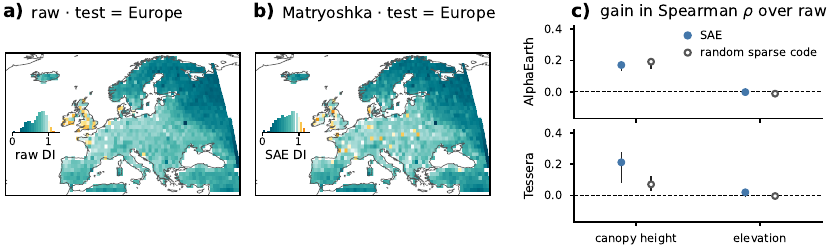}
    \caption{Area of applicability for canopy height (CHMv2~\cite{brandt.etal_2026}) with Europe held out.
    \textbf{(a, b)}~Dissimilarity index of every held-out pixel, divided by the threshold of Appendix~\ref{sec:supp_aoa} so that $1$ marks the boundary of the area of applicability, in the raw AlphaEarth embedding and in the Matryoshka SAE code; the index uses the ridge-weighted distance to the $5$ nearest training pixels.
    \textbf{(c)}~Gain in Spearman $\rho$ of the ridge-weighted distance in a sparse code over the raw distance in ranking the errors of one ridge head fitted on the raw embedding, on graded hold-outs (Appendix~\ref{sec:supp_aoa_graded}; median and $95\%$ bootstrap interval over folds), for a BatchTopK SAE (three seeds) and for an untrained random sparse code, on AlphaEarth and Tessera; the same panel is Figure~\ref{fig:aoa_main}.
    The gain is specific to the target: positive for canopy height, absent for elevation.}
    \label{fig:aoa}
\end{figure}

\paragraph{Details of Figure~\ref{fig:aoa_main}.} Each point is the median over the $88$ graded hold-outs of Appendix~\ref{sec:supp_aoa} of the paired gain in Spearman $\rho$ of the ridge-weighted distance in a sparse code over the raw distance; the bars are $95\%$ fold-bootstrap intervals.
For the BatchTopK SAE ($M{=}1024$, $k{=}16$), the gain of each fold is the mean over three seeds.
The random sparse code is untrained, has the same dictionary size and sparsity, and is weighted by the coefficients of its own ridge head.
The panel is panel c of Figure~\ref{fig:aoa}.

\begin{figure}[ht]
    \centering
    \includegraphics[width=\linewidth]{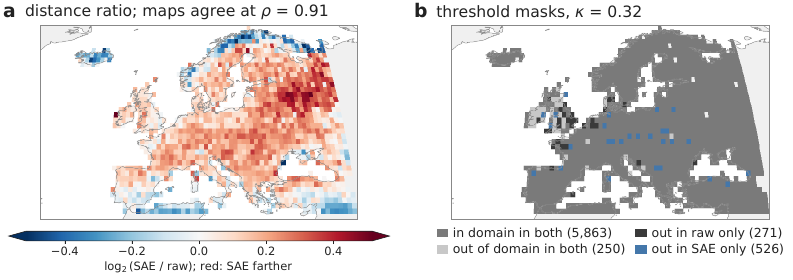}
    \caption{Area of applicability for canopy height with Europe held out ($6{,}910$ held-out pixels), from the ridge-weighted distance to the $5$ nearest training pixels in the raw embedding and in the Matryoshka SAE code.
    \textbf{(a)}~$\log_2$ ratio of the SAE and raw distances; positive values mark pixels the SAE places relatively farther from the training data.
    \textbf{(b)}~Pixels outside the area of applicability in each of the two spaces, overlaid.}
    \label{fig:supp_aoa_diff}
\end{figure}

\subsection{The leave-Europe-out fold}\label{sec:supp_aoa_europe}

The fold mapped in Figures~\ref{fig:aoa} and~\ref{fig:supp_aoa_diff} holds out Europe, with canopy height as the target: the head is trained on the five other continents and applied to Europe.
On this fold the raw distance is slightly anti-correlated with the head's error, and the SAE distance is positively correlated on all four checkpoints (Table~\ref{tab:supp_aoa_europe}).
The two heads predict equally well here (test $R^2$ $0.676$ raw, $0.658$--$0.675$ SAE), so the difference is not an artefact of a weaker head.
The gain in Spearman $\rho$ of the SAE distance over the raw distance depends on the checkpoint: $+0.16$ on Matryoshka, which Figures~\ref{fig:aoa} and~\ref{fig:supp_aoa_diff} map, and $+0.30$ to $+0.44$ on the three BatchTopK seeds.

\begin{table}[h]
\centering
\small
\caption{Spearman $\rho$ between the ridge-weighted distance and the absolute error of the canopy-height head, where each distance is evaluated against the head fitted on the same representation, on the train-on-Europe and leave-Europe-out folds.
The raw-embedding values do not depend on the SAE checkpoint.}
\label{tab:supp_aoa_europe}
\begin{tabular}{lcccc}
\toprule
 & \multicolumn{2}{c}{train on Europe} & \multicolumn{2}{c}{leave Europe out} \\
\cmidrule(lr){2-3}\cmidrule(lr){4-5}
checkpoint & raw & SAE & raw & SAE \\
\midrule
BatchTopK, seed 42 & $+0.44$ & $+0.45$ & $-0.04$ & $+0.36$ \\
BatchTopK, seed 43 & $+0.44$ & $+0.51$ & $-0.04$ & $+0.26$ \\
BatchTopK, seed 44 & $+0.44$ & $+0.48$ & $-0.04$ & $+0.40$ \\
Matryoshka         & $+0.44$ & $+0.42$ & $-0.04$ & $+0.12$ \\
\bottomrule
\end{tabular}
\end{table}

The SAE distance changes the extremes of the map, not its overall pattern.
The two distance maps are rank-correlated at Spearman $0.91$ over the held-out pixels.
Most of Europe lies inside the area of applicability, $92.5\,\%$ in raw space and $88.8\,\%$ in SAE space.
The two spaces disagree on whether a pixel lies inside the area of applicability for $797$ of $6{,}910$ pixels ($11.5\,\%$; Cohen's $\kappa = 0.32$), and $526$ of them lie outside it in SAE space only (Figure~\ref{fig:supp_aoa_diff}).

On the opposite fold, train-on-Europe, the two spaces give nearly the same map: their unweighted distances rank the held-out pixels at Spearman $0.94$ to $0.96$.
The raw distance already ranks the error well there (ridge-weighted $\rho = +0.44$), which leaves a sparse code little to add.
There the SAE heads also predict worse than the raw head (test $R^2$ $0.14$ to $0.24$ against $0.34$), and larger errors are easier to rank.
The slightly higher SAE correlations of Table~\ref{tab:supp_aoa_europe} on that fold are therefore not evidence for the sparse code.

\subsection{Continent and PFT folds}\label{sec:supp_aoa_folds}

On the continent and plant-functional-type (PFT) folds of Appendix~\ref{sec:supp_aoa}, each distance is scored by its Spearman correlation with the absolute error of the head fitted on the same representation.
Over all $208$ cells, the ridge-weighted SAE distance ranks the error better than the raw distance in $165$ ($79\,\%$), with a median gain of $+0.10$.
The gain holds on each checkpoint: $44$, $39$ and $40$ of $52$ cells for the three BatchTopK seeds, $42$ of $52$ for Matryoshka.
The unweighted distance gains less, in $140$ of $208$ cells ($67\,\%$, median $+0.05$), and on the Matryoshka checkpoint in only $29$ of $52$.
Without ridge weights, the SAE distance is thus not consistently better on both SAE variants.
The gain concentrates where the raw distance fails: the SAE is better in $95$ of $104$ leave-one-out cells (median $+0.15$) and in $70$ of $104$ single-region cells (median $+0.06$).
Canopy height gains more than elevation, $95$ of $104$ cells (median $+0.18$) against $70$ of $104$ ($+0.05$).

\subsection{The gain is specific to the target}\label{sec:supp_aoa_graded}

Scoring each distance against its own head mixes two effects: a better distance and a different head.
To separate them, we fit one ridge head on the raw embedding and score every distance by how well it ranks the errors of that head, as in §\ref{sec:results_aoa}.
These hold-outs are graded by how far they lie from the training data (Appendix~\ref{sec:supp_aoa}), and the score is the gain in Spearman $\rho$ over the raw distance.

With the head fixed, the ridge-weighted distance in a BatchTopK code ($M{=}1024$, $k{=}16$, three seeds) ranks canopy-height error better than the raw distance in $77$ of $88$ folds on AlphaEarth (median gain $+0.17$) and in $75$ of $88$ on Tessera ($+0.21$; Figure~\ref{fig:aoa_main}).
For elevation the gain is absent: the SAE distance is better in $41$ of $88$ folds on AlphaEarth (median $0.00$) and in $49$ of $88$ on Tessera ($+0.02$).
The gain for elevation on the continent and PFT folds above thus comes mostly from the head: the SAE head fits elevation worse than the raw head, and its larger errors are easier to rank.

Two further observations limit how the canopy-height gain can be interpreted.
First, it benefits from the head's weights: the unweighted distance gains less on the continent and PFT folds.
Second, on AlphaEarth it does not need a learned dictionary: an untrained random sparse code of the same dictionary size and sparsity, weighted by its own head in the same way, gains as much ($75$ of $88$ folds, median $+0.19$).
On Tessera this random code gains about a third as much ($57$ of $88$, $+0.07$).
What helps is therefore sparsity combined with the head's weights, which concentrate the distance on the few features the head relies on.

Why would this help for canopy height and not for elevation?
We can only offer an interpretation.
Canopy height is tied to vegetation type and structure, which a sparse code represents with a few features that are active on some land-cover types and silent on others.
A head-weighted distance in such a code is large when a test pixel activates features that are never active in the training region.
Elevation varies continuously within every land-cover type and is spread over many embedding dimensions at once, so no small set of sparse features marks a pixel as novel for this target.
Sparsifying the representation cannot add a signal that the embedding does not contain in that form.

\subsection{Feature scaling and the choice of weights}\label{sec:supp_aoa_scaling}

In the results above, features are not rescaled before the distance is computed.
We test two rescalings, applied to the features of both spaces: dividing each feature by its standard deviation over the training pixels of the fold, and dividing it by its mean magnitude over the training pixels on which it is non-zero.
Both reduce the advantage of the unweighted SAE distance, which stays ahead of the raw distance in $121$ and $133$ of the $208$ continent and PFT cells, against $140$ unscaled.
Under the first rescaling, its median gain in Spearman $\rho$ falls from $+0.05$ to $+0.02$.
With ridge weights, the SAE distance stays ahead under either rescaling, in $180$ and $169$ cells against $165$ unscaled.
Weighting features by random-forest importance in place of ridge coefficients, as in the original recipe of Meyer and Pebesma~\cite{meyer.pebesma_2021}, leaves the SAE ahead in $129$ of $208$ cells.

\subsection{Implications}\label{sec:supp_aoa_implications}

Beyond the reasons given in §\ref{sec:results_aoa}, the results of this appendix have further implications for how sparse applicability estimates could be used and evaluated.

\paragraph{Use the sparse distance next to the raw one.} The sparse distance helps most where the raw distance fails to rank the error, as on the leave-Europe-out fold, and adds little where the raw distance already works.
On the train-on-Europe fold, the ridge-weighted raw distance reaches $\rho=+0.44$, and under the unweighted distance the two spaces flag nearly the same pixels as outside the area of applicability ($61\,\%$ and $60\,\%$, Cohen's $\kappa=0.94$).
A practitioner does not know in advance which case applies.
Computing both estimates costs little, and the pixels on which they disagree, $11.5\,\%$ of Europe in Figure~\ref{fig:supp_aoa_diff}, are a natural list of places to check first, for instance with field data.

\paragraph{A better ranking is not yet a better map.} The Spearman $\rho$ rates how well a distance orders the pixels by their error, whereas a map of the area of applicability also needs a threshold.
The threshold of Meyer and Pebesma is calibrated by spatial cross-validation on the training pixels, separately in each space, so the same rule excludes different shares of pixels in the two spaces ($7.5\,\%$ and $11.2\,\%$ of Europe).
Whether either threshold separates large from small errors well is a question that the rank correlation does not answer.
The correlations are also modest, with a median $\rho$ of $+0.32$ for the SAE distance over the $208$ continent and PFT cells.
An applicability map thus says which areas deserve scrutiny, not where a model stops being valid.

\paragraph{Keep the prediction model, change only the distance.} Fitting the downstream model on the sparse code can make its predictions worse: on the train-on-Europe fold, the SAE heads reach a test $R^2$ of $0.14$ to $0.24$, against $0.34$ for the raw head.
This is not needed, because on the graded hold-outs the sparse distance ranks the errors of a head fitted on the raw embedding.
Only the weights of the distance come from a head fitted on the sparse code, and they matter.
Without them the gain shrinks, and with random-forest importance in place of ridge coefficients, the SAE distance is ahead in only $129$ instead of $165$ of the $208$ cells (Appendix~\ref{sec:supp_aoa_scaling}).
The estimate therefore depends on choices beyond the representation, namely the weights and the scaling of the features, and a study should report them with it.

\paragraph{When to expect a gain.} Our interpretation of why canopy height gains and elevation does not (Appendix~\ref{sec:supp_aoa_graded}) suggests a condition that could be checked before an estimate is used.
The sparse distance should help for targets that are tied to landscape types, which a sparse code marks with a few features, and not for targets that vary continuously within every type.
If this holds, the share of the head's weight that its few largest features carry could indicate in advance whether the sparse distance helps; we have not tested this.

\paragraph{A weak proxy for concept shift.} A distance cannot detect concept shift (§\ref{sec:results_aoa}), but the two kinds of shift often come together.
A land-cover type that is absent from training tends to bring a relation to the target that the head has not seen.
In so far as sparse features mark such types, the sparse distance is also a weak proxy for concept shift, which only labels from the new region can confirm.

\paragraph{Comparing applicability estimates.} Three lessons concern evaluation.
First, every distance should be scored against the errors of one common head.
Scored against its own head, the SAE distance appeared to help for elevation ($70$ of $104$ continent and PFT cells), but this gain came mostly from a weaker head, whose larger errors are easier to rank, and it disappeared with a common head.
Second, one target is not enough, since canopy height and elevation lead to opposite conclusions on the same hold-outs.
Third, hold-outs built around shared centres are not independent, so the uncertainty of a gain should be estimated over centres rather than over folds.

\paragraph{Towards interpretable applicability maps.} The weighted distance in a sparse code is a sum over features, to which only the features active on the test pixel or on its training neighbours contribute.
For a pixel outside the area of applicability, one can therefore list the few features that place it far from the training data.
If these features were named concepts, the map would say which concept is missing from training, not only that the pixel is far from it.
The current SAEs do not allow this yet: most of their features are aligned with no concept group (§\ref{sec:results_emf}), and on AlphaEarth a random sparse code gains as much, so the gain does not rest on interpretable features.
The mechanism that motivated the approach, that polysemantic dimensions create spurious neighbours, also remains untested.
A direct test would hold out a concept that shares a polysemantic raw dimension with a concept of the training region, chosen with the alignment heatmaps of Appendix~\ref{sec:supp_alignment}, and ask whether the sparse distance flags it where the raw distance does not.
SAE architectures and losses tailored to Earth embeddings could yield features closer to the physical drivers of a target, and these are the features that an interpretable applicability map needs.

\section{Broader Impacts}\label{sec:broader_impacts}

More interpretable Earth embeddings benefit downstream users in agriculture, ecology, hydrology, and disaster response by making clear what an upstream foundation model has and has not encoded.
For some targets, they also give practitioners a sharper way to flag locations where a downstream model is extrapolating.
Together, these address one of the main barriers to trustworthy global mapping.
The negative consideration is that finer-grained, globally indexed feature dictionaries could in principle support surveillance applications (e.g.\ identifying specific land-use or settlement patterns at scale).
We work on field-scale ($10$\,m) public embeddings, release code, an evaluation pipeline and a dataset recipe rather than identifying labels, and report results in aggregate rather than for individual sites.
We do not see a clear additional misuse vector beyond what the upstream embeddings already enable.

\FloatBarrier
\clearpage

\newpage
\section*{NeurIPS Paper Checklist}

\begin{enumerate}

\item {\bf Claims}
    \item[] Question: Do the main claims made in the abstract and introduction accurately reflect the paper's contributions and scope?
    \item[] Answer: \answerYes{}
    \item[] Justification: The contributions stated in §\ref{sec:intro} (first SAE application to pixelwise Earth embeddings, evaluation protocol with reconstruction at matched sparsity, concept alignment and spatial scale, each compared with baselines that include a frozen-decoder SAE, a comparison of three SAE variants, and two illustrated applications) are each reported with the corresponding experimental evidence in §\ref{sec:results} and the supplement. The abstract and introduction state which scores separate a trained SAE from the frozen-decoder baseline (reconstruction) and which do not (concept alignment, spatial scale), and that the applicability benefit depends on the target.

\item {\bf Limitations}
    \item[] Question: Does the paper discuss the limitations of the work performed by the authors?
    \item[] Answer: \answerYes{}
    \item[] Justification: The Discussion states the limitations: statistics computed from single-feature activations, such as concept-alignment counts and spatial scales, do not separate a trained SAE from a frozen random decoder, many atoms are regional rather than concept-pure, and the area-of-applicability gain is target-dependent (canopy height but not elevation) and addresses covariate shift only. Appendix~\ref{sec:supp_aoa_results} and Appendix~\ref{sec:supp_subconcepts} give the negative results in detail.

\item {\bf Theory assumptions and proofs}
    \item[] Question: For each theoretical result, does the paper provide the full set of assumptions and a complete (and correct) proof?
    \item[] Answer: \answerNA{}
    \item[] Justification: The paper is empirical. No formal theorems are stated and no proofs are required.

\item {\bf Experimental result reproducibility}
    \item[] Question: Does the paper fully disclose all the information needed to reproduce the main experimental results of the paper to the extent that it affects the main claims and/or conclusions of the paper (regardless of whether the code and data are provided or not)?
    \item[] Answer: \answerYes{}
    \item[] Justification: Architecture, training schedule, optimisation hyperparameters, dataset construction, and evaluation samples are documented in §\ref{sec:data} and Appendix~\ref{sec:supp_impl}. Embeddings (AlphaEarth, Tessera) and reference products are public.

\item {\bf Open access to data and code}
    \item[] Question: Does the paper provide open access to the data and code, with sufficient instructions to faithfully reproduce the main experimental results, as described in supplemental material?
    \item[] Answer: \answerYes{}
    \item[] Justification: The code (dataset construction, training, inference, evaluation) and the dataset recipe are released at \url{https://github.com/EarthyScience/geosae}. All input embeddings (AlphaEarth, Tessera) and reference products are publicly available; the MajorTOM grid we sample from is also public.

\item {\bf Experimental setting/details}
    \item[] Question: Does the paper specify all the training and test details (e.g., data splits, hyperparameters, how they were chosen, type of optimizer) necessary to understand the results?
    \item[] Answer: \answerYes{}
    \item[] Justification: §\ref{sec:supp_data}, §\ref{sec:supp_eval}, and §\ref{sec:supp_training} document the train/val/test split, optimiser (Adam), schedule (cosine annealing with warmup), all SAE hyperparameters, and the sampling protocol for the evaluation universes.

\item {\bf Experiment statistical significance}
    \item[] Question: Does the paper report error bars suitably and correctly defined or other appropriate information about the statistical significance of the experiments?
    \item[] Answer: \answerYes{}
    \item[] Justification: The sweep of Figure~\ref{fig:pareto} is run with three seeds per cell on both products, and the frozen-decoder baseline with three seeds. Per-feature statistics in §\ref{sec:results_freq} and §\ref{sec:results_emf} are shown as distributions over all features rather than point estimates. Results over folds in §\ref{sec:results_aoa} and Appendix~\ref{sec:supp_aoa_results} are reported as counts of folds and medians. The held-out evaluation slice is fixed at $100$k patches.

\item {\bf Experiments compute resources}
    \item[] Question: For each experiment, does the paper provide sufficient information on the computer resources (type of compute workers, memory, time of execution) needed to reproduce the experiments?
    \item[] Answer: \answerYes{}
    \item[] Justification: Appendix~\ref{sec:supp_training} reports the hardware (a single NVIDIA A100 or A40 GPU per run), the logged training time per run (median $22$\,min on AlphaEarth and $58$\,min on Tessera) and the total of the sweep of Figure~\ref{fig:pareto} ($108$ runs, $73$ GPU-hours). The project required preliminary runs and evaluation passes whose cost was not logged and is not included in this total.

\item {\bf Code of ethics}
    \item[] Question: Does the research conducted in the paper conform, in every respect, with the NeurIPS Code of Ethics?
    \item[] Answer: \answerYes{}
    \item[] Justification: The work uses public Earth observation embeddings and public categorical/continuous reference products, with no human subjects, no scraped personal data, and no dual-use risk beyond the broader debate about geospatial monitoring discussed under Broader Impacts.

\item {\bf Broader impacts}
    \item[] Question: Does the paper discuss both potential positive societal impacts and negative societal impacts of the work performed?
    \item[] Answer: \answerYes{}
    \item[] Justification: Appendix~\ref{sec:broader_impacts} (``Broader Impacts'') addresses both positive applications (more interpretable, trustworthy global mapping) and negative considerations (finer-grained dictionaries could in principle support surveillance), with the mitigations we already apply by working at $10$\,m resolution and releasing code and a dataset recipe rather than identifying labels.

\item {\bf Safeguards}
    \item[] Question: Does the paper describe safeguards that have been put in place for responsible release of data or models that have a high risk for misuse?
    \item[] Answer: \answerNA{}
    \item[] Justification: The released artefacts are code, an evaluation pipeline and a dataset recipe that operate on publicly available Earth embeddings. None carries a high misuse risk beyond what the upstream embeddings already pose.

\item {\bf Licenses for existing assets}
    \item[] Question: Are the creators or original owners of assets used in the paper properly credited and are the license and terms of use explicitly mentioned and properly respected?
    \item[] Answer: \answerYes{}
    \item[] Justification: All embedding products and reference products are cited at first use in §\ref{sec:data}. The Esri World Imagery basemap used for qualitative figures and the attribution statements required by the data providers are given in the Acknowledgments.

\item {\bf New assets}
    \item[] Question: Are new assets introduced in the paper well documented and is the documentation provided alongside the assets?
    \item[] Answer: \answerYes{}
    \item[] Justification: The code, evaluation pipeline, concept registry and dataset recipe (MajorTOM cell identifiers, splits and reference-product preprocessing) are released with documentation describing training configuration, evaluation protocol, and licensing.

\item {\bf Crowdsourcing and research with human subjects}
    \item[] Question: For crowdsourcing experiments and research with human subjects, does the paper include the full text of instructions given to participants and screenshots, if applicable, as well as details about compensation (if any)?
    \item[] Answer: \answerNA{}
    \item[] Justification: The paper does not involve crowdsourcing or research with human subjects.

\item {\bf Institutional review board (IRB) approvals or equivalent for research with human subjects}
    \item[] Question: Does the paper describe potential risks incurred by study participants, whether such risks were disclosed to the subjects, and whether Institutional Review Board (IRB) approvals were obtained?
    \item[] Answer: \answerNA{}
    \item[] Justification: The paper does not involve research with human subjects.

\item {\bf Declaration of LLM usage}
    \item[] Question: Does the paper describe the usage of LLMs if it is an important, original, or non-standard component of the core methods in this research?
    \item[] Answer: \answerNA{}
    \item[] Justification: LLMs were used only as a writing aid (drafting and editing) and play no role in the methodology, results, or scientific rigor of the work.

\end{enumerate}

\end{document}